\documentclass{article}
\usepackage[utf8]{inputenc}
\usepackage[english]{babel}

\usepackage{lmodern}

\usepackage{mystyle}

\newtheorem{theorem}{Theorem}[section]

\theoremstyle{definition}
\newtheorem{definition}[theorem]{Definition}

\AtEveryBibitem{\clearfield{month}}
\AtEveryBibitem{\clearfield{doi}}
\AtEveryBibitem{\clearfield{volume}}
\AtEveryBibitem{\clearfield{eprinttype}}
\AtEveryBibitem{\clearfield{eprint}}

\usepackage{ifthen}
\newboolean{is_thesis}

\begin{document}
\setboolean{is_thesis}{false}

\title{The Topological BERT:\\ Transforming Attention into Topology\\ for Natural Language Processing}

\newcommand*\samethanks[1][\value{footnote}]{\footnotemark[#1]}
\author{Ilan Perez\thanks{Ecole Polytechnique Fédérale de Lausanne (EPFL), Laboratory for topology and neuroscience,CH-1015 Lausanne, Switzerland} (\href{mailto:ilan.perez@epfl.ch}{ilan.perez@epfl.ch}),\\ Raphael Reinauer\samethanks[1] (\href{mailto:raphael.reinauer@epfl.ch}{raphael.reinauer@epfl.ch})}

\maketitle

\begin{abstract}
\ifthenelse{\boolean{is_thesis}}{
    \begin{center}
        \textbf{Abstract}
    \end{center}
    }{}
    In recent years, the introduction of the Transformer models sparked a revolution in natural language processing (NLP).
    BERT was one of the first text encoders using only the attention mechanism without any recurrent parts to achieve state-of-the-art results on many NLP tasks. 

    This paper introduces a text classifier using topological data analysis.
    We use BERT's attention maps transformed into attention graphs as the only input to that classifier. 
    The model can solve tasks such as distinguishing spam from ham messages, recognizing whether a sentence is grammatically correct, or evaluating a movie review as negative or positive.
    It performs comparably to the BERT baseline and outperforms it on some tasks. 
    
    Additionally, we propose a new method to reduce the number of BERT’s attention heads considered by the topological
classifier, which allows us to prune the number of heads from 144 down to as few as ten with no reduction in performance.
    Our work also shows that the topological model displays higher robustness against adversarial attacks than the original BERT model, which is maintained during the pruning process.  To the best of our knowledge, this work is the first to confront topological-based models with adversarial attacks in the context of NLP.
\end{abstract}



\section*{Introduction}
\ifthenelse{\boolean{is_thesis}}{\enlargethispage{10\baselineskip}}{}
In 2003, Yoshua Bengio and his team proposed the first neural network for natural language processing (NLP) (\cite{first_neural_net_nlp}). 
Since then, there has been a festival of new model architectures surpassing current records on many textual tasks. 
Recurrent neural networks (RNNs) (\cite{first_recurrent_neural_net}) were replaced by long short-term memory (LSTM) networks (\cite{first_long_short_term_memory_net}) before attention-based models were proposed (\cite{transformer}).

One of the first text encoder models using only the attention mechanism without any recurrent parts is BERT, and it got state-of-the-art performance on many NLP benchmark tasks. 
But its success came with the desire to understand what type of language knowledge these models acquire, such as grammar rules, semantics, syntactic relations between words, or even the world knowledge it can infer through language. 
Numerous studies about these different subjects are very well summarized under the term Bertology (\cite{bertology}).

The geometric and topological information contained in BERT has recently caught the interest of the topological data analysis (TDA) community (\cite{introduction_persistent_homology_comp_science, acceptability_judgement_topo_att_maps, artificial_text_topo_att_maps}).
The attention head activation can be transformed into an attention graph. 
One can then filter this graph and apply persistent homology to study the evolution of connected components and higher-order structures, compute a collection of Betti numbers, or store the average barcode length of a given homology dimension.
This approach provides a topological representation of textual input and can be used to train a new type of textual classifier or find new interpretability methods for NLP.

The contributions of this paper are summarized as follows:
\begin{enumerate}[label=(\roman*)]

    \item We reproduce the results of \cite{betti_bert} but in a broader setting with different choices of attention graph filtrations, types of persistence homology (ordinary and directed), and symmetry functions. In addition, we also use different topological features (they considered Betti numbers, and we consider persistence images).
    
    \item We study UMAP projections of persistent diagrams obtained via the attention graphs and compare their distribution with the classification of attention maps proposed by \cite{dark_secret_of_bert}. 
    We could not observe any correlation, but we find stability across input sentences of the cluster composition, indicating a specialization of the attention heads. 
    
    \item We propose a new method to rate attention heads inspired by GradCam (\cite{gradcam}) on the topological inputs that works remarkably well to prune the number of heads down from 144 to ten with no reduction either in classification performance or robustness against adversarial attacks.
    Moreover, the selected heads largely display attention patterns focusing on the [SEP] token, leading to a new perspective on the \textit{no-op} hypothesis proposed in \cite{what_does_bert_look_at}.
    
    \item Our research is the first study that contemplates the topic of adversarial attacks in relation to topological-based models for NLP tasks.
\end{enumerate}

\ifthenelse{\boolean{is_thesis}}{
The code of the paper can be found at \href{https://github.com/raphaelreinauer/PersistenceAttention}{https://github.com/raphaelreinauer/PersistenceAttention}.
}{}

\section{Related Work}

\textbf{Topological Data Analysis} Combining algebraic topology and machine learning has become a vast field of investigation in the past decade.
To the best of our knowledge, \cite{introduction_persistent_homology_comp_science} was the first to use persistent homology in the context of NLP. 
The author differentiated child writing from teenager writing using persistence tools. 
This is followed by an increase in interest from the scientific community in TDA methods in NLP, including a successful attempt to predict the genre of a movie from its plot (\cite{tda_movie_genre}), an application of persistent homology to depict textual entailment in legal processes (\cite{tda_legal_entailment}), an unsuccessful TED-talk rating (\cite{tda_tedtalk}), and detection of artificially generated text (\cite{artificial_text_topo_att_maps}).

To demonstrate this increase in interest, \cite{acceptability_judgement_topo_att_maps} examined, independently and in parallel to our work, the linguistic acceptability of text using the topology of the attention graphs. 
They were able to enhance the performance of existing models for two standard practices in linguistics: binary judgments and linguistic minimal pairs.
They also proposed a method for analyzing the linguistic functions of attention heads and interpreting the correspondence between the graph features and grammatical phenomena.


The challenge of combining a neural network with a topological layer is the differentiability of the overall objective function. 
The PersLay model (\cite{perslay}) obtained excellent classification performance of real-life graph datasets such as social graphs or data from medical or biological domains. This is similar to Persformer (\cite{persformer}), which can process persistent diagrams without using handcrafted features but using the self-attention mechanism and achieving state-of-the-art results on the standard ORBIT5K dataset.


\textbf{BERT Model} Multiple studies have shown that BERT is overparametrized. 
\cite{dark_secret_of_bert} obtained an increase in performance of the model when disabling the attention in certain attention heads.
\cite{are_sixteen_heads_better_than_one} proposed a pruning method that disabled 40\% of the attention heads while keeping the accuracy high and the authors showed that some layers could be reduced to one attention head for better performance. 
Even the original transformer architecture has been pruned from 48 heads down to 10 heads while maintaining the accuracy level (\cite{analysing_multi_head_attention}). 
Interestingly. these remaining heads displayed specific interpretable linguistic functions. 
These specialized heads have also been found in BERT by \cite{what_does_bert_look_at}, who present attention heads that attend to the direct objects of verbs, to the determiner of nouns, or to coreferent mentions (\cite{what_does_bert_look_at}). 

\cite{bert_plays_lottery} showed that BERT contained many subnetworks achieving performance comparable to the full model, and furthermore, that the worst performing subnetwork remains highly trainable. 

Another approach to illustrate the information contained in a pretrained model is the study of the transferability of the contextual representations stored inside the model. 
\cite{linguistic_knowledge_and_transferability_of:contextual_representations} found that linear models trained on top of frozen contextual representations, such as the pretrained BERT model, are competitive with state-of-the-art task-specific models. 
This is related to our work, as we also extract information from a frozen BERT model and train a topological classifier on it.

\section{Background}
\subsection{BERT Model}
The multi-headed attention model BERT (\cite{bert}) was one of the first models to use only the encoder part of the original transformer architecture (\cite{transformer}) while obtaining state-of-the-art results on many NLP tasks. 
It is a model composed of $L$ encoder layers connected in series each containing $H$ attention heads applied in parallel. 

We focus on the attention heads of $\text{BERT}_\text{BASE}$ ($L = 12, H=12$) as they are the part of the model containing the topological information we want to analyse. 
The input of an attention head is a matrix $X$ of size $n\times d$ (where $d=512$) whose rows are vector representations of the $n$ tokens of the input sentence, and the output is a new representation $X^\mathrm{out}$ of size $n \times d'$ with $d' < d$ ($d' = 64$) following the formula:
\begin{equation}
    X^\mathrm{out} = W^\mathrm{attn} \cdot (X \cdot W^V) \\
    \text{ with } W^\mathrm{attn} = \text{softmax} \bigg( \frac{(X \cdot W^Q)\cdot (X \cdot W^K)^T}{\sqrt{d}} \bigg)
\end{equation}
where $W^Q,\, W^K,\, W^V \in \mathbb{R}^{d\times d'}$ are trainable matrices that project the token embeddings into a lower-dimensional vector space, and the softmax is token over the second dimension. 
The matrix $W^\mathrm{attn}$ is a $n \times n$-dimensional matrix called the attention matrix.

Its entries $w_{ij}^\mathrm{attn}$ can be interpreted as the attention given to the token at position $j$ when computing the new representation of the token at position $i$. 
They are non-negative and the attention scores of a token sum up to one, i.e., $\sum_{j=1}^n w_{ij}^\mathrm{attn}=1$ for all $i=1,\ldots, n$.

\subsection{Attention Maps and Attention Graphs}
\begin{figure}[ht]
  \centering
  \includegraphics[scale=0.5]{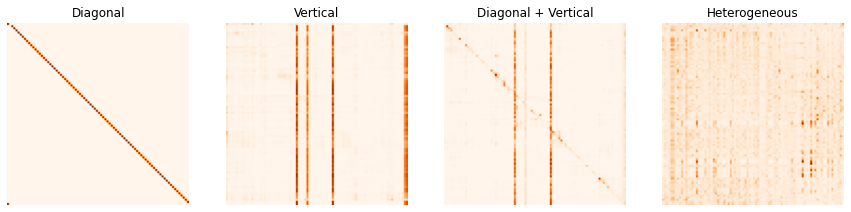}     
  \caption {Typical self-attention maps depicted by \cite{dark_secret_of_bert}. Both axes on every attention map represent BERT tokens of an input sentence. The colors denote attention weights: the darker the color, the larger the weight. The authors assumed that only the last two types potentially encode syntactic and semantic information, while the first three are associated with model pretraining.}
  \label{fig:self_attention_map_patterns}
\end{figure}

Attention maps are the representations of the attention matrices into $d\times d$ pixel images. 
The higher $w_{ij}^{attn}$ is, the darker the pixel color is. 
Those maps where intensively studied in \cite{dark_secret_of_bert} and divided into four classes: diagonal patterns, vertical patterns, diagonal and vertical patterns, and heterogeneous patterns (see \cref{fig:self_attention_map_patterns}). 

Another representation of an attention matrix is through an attention graph. 
Given a head, we construct a weighted directed graph taking as vertices the tokens of the input sentence and connecting two words $i$ and $j$ with an edge from $i$ to $j$ with weight $w_{ij}^\mathrm{attn}$ and an opposite direction edge with weight $w_{ji}^\mathrm{attn}$. 
No further modification is needed to apply directed persistent homology and in the non-directed versions, this directed graph is transformed into a complete non-directed weighted graph on the set of tokens via a symmetric function $f$. 
The edge connecting tokens $i$ and $j$ is assigned to the weight $1 - f(w_{ij}^\mathrm{attn}, w_{ji}^\mathrm{attn})$. 
The larger the weight of the edge connecting the two vertices, the lower the transformed attention score between the two tokens. 
The transformation from attention map to attention graph is illustrated in \cref{fig:transformation_procedure_attention_maps_to_persistence_images}.

\subsection{Persistent Homology}\label{sec:persistent_homology}
The attention graphs constructed from the attention heads contain the topological structure we are interested in.
Topologically, a weighted graph and the corresponding unweighted graph are identical. 
To encode the topological information provided by the graph weights, we use filtrations of the graph. 
A filtration is a sequence of nested topological subspaces indexed either on a discrete set like $\{1, \ldots, n \}$ or on a continuous real-valued parameter. 
Starting with an attention graph, we consider three types of filtrations: Ordinary, MultiDim, and Directed.

\begin{description}[wide,itemindent=\labelsep]
    \item[Ordinary] We initiate the filtration with only the vertices of the graph. 
    Then we add edges one by one, depending on their weights, until we obtain the complete graph (see \cref{fig:transformation_procedure_attention_maps_to_persistence_images}). 
    The order of how the edges are added is as follows:
    Given two edges $e_1$ and $e_2$, if the weight of $e_1$ is smaller than the weight of $e_2$, then $e_1$ will be added before $e_2$. 
    This filtration is based on a real-valued parameter $t$ taking value in $[0,1]$ and the filtration at stage $t_0$ is formed by all the edges with weights smaller or equal to $t_0$.
    
    \item[MultiDim] The Ordinary filtration can be seen as starting with 0-simplices and then adding 1-simplices to construct the graph. 
    The 0-simplices can be thought of as points, the 1-simplices as edges, the 2-simplices as triangles, these 3-simplices as tetrahedrons, and so on.
    The edges for the MultiDim filtration have the same filtration values as for the Ordinary filtration, but we add a 2-simplex everytime three edges form a triangle. 
    
    \item[Directed] We consider the directed version of the attention graph and again start the filtration with only the vertices. 
    The idea is similar to the MultiDim filtration: 
    we add edges one by one, depending on their weight, and we add a 2-simplex if its boundary 1-simplices are present and do not form a directed cycle.
\end{description}

Filtrations are the topological interpretation of edge weights and their directions, for if the weights on the edges or the direction of the edges were different, the filtration would change accordingly.

Given a filtered simplicial complex, we can analyze it through persistent homology.
The idea is to keep track of the appearance and disappearance of topological features through the filtration by computing the homology of each topological space encountered during the filtration and keeping track of the maps induced by the inclusions. 
An introduction to the mathematical background of TDA can be found in \cite{introduction_persistent_homology_comp_science}.
One can think of 0-dimensional persistent features as connected components, 1-dimensional features as holes and 2-dimensional features as cavities (2-dimensional holes). 
The birth time of a persistent feature is the filtration value at which the feature appears.
For examples, the birth time of all the 0-dimensional features of our graph filtration is 0 and the birth time of a 1-dimensional features is the filtration value of the edge completing a graph cycle. 
The death time is the filtration value at which the feature disappears.
For a 0-dimensional feature it will correspond to the weight of the edge connecting the corresponding vertex to the main connected component. 
If there are no 2-simplices in the filtration, a 1-dimensional feature will never vanish, and its death time is said to be equal to $\infty$. But for computational purpose, it is set to the maximal filtration value $1.0$.
The birth and death times of each feature are stored in a persistence diagram. A persistent feature is seen as a point in $\mathbb{R}^2$ with its birth time as $x$-coordinate and death time as $y$-coordinate (see \cref{fig:transformation_procedure_attention_maps_to_persistence_images}). 
Hence a persistence diagram is a multi-set of elements in $\mathbb{R}^2$ -- \emph{multi} because two or more persistent features may have the same birth and death time.

\subsection{Persistence Images}
Incorporating persistence diagrams into a machine learning pipeline faces two main challenges. 
Firstly, one can add distances on the space of persistence diagrams, such as the bottleneck distance or a Wasserstein-type distance (see \cref{def:wasserstein_distances}). 
But the underlying data structure remains a set, not a vector. 
Secondly, the space of persistence diagrams cannot be isometrically embedded into a Hilbert space, and this remains true for any type of distance considered (\cite[Theorem 4.3]{persistence_diagrams_not_hilbert_space}).

To overcome this issue, the TDA community has developed many methods to convert a collection of persistence diagrams into a collection of vectors contained in a Hilbert space. 
The method we consider uses persistence images (\cite{persistence_images}) because they are simple to compute and interpret. Furthermore, they are compatible with convolutional neural networks. 

A persistence diagram $\mathcal{D}$ can be seen as a non-continuous map $f\colon \mathbb{R}^2 \to \mathbb{N}$ that counts the number of points in the diagram at the input location $u = (x,y)$. 
This function can be decomposed into a finite sum of indicator functions $f_{u}$ that return 1 if the input is $u$ and 0 otherwise. 
These indicator functions can be seen as probability density functions. 
To work with continuous functions, $f_{u}$ will be approximated by the 2-dimensional Gaussian distribution $\phi_{u}$ of of mean $u$ and standard deviation $\sigma$, a hyperparameter that has to be chosen. 
We sum up all those continuous functions to obtain $g$, a continuous approximation of the function $f$. 

For more flexibility and stability, a weight function $w\colon \mathbb{R}^2 \to \mathbb{R}_{\geq 0}$ is incorporated inside the sum to emphasize certain regions of the persistence diagram. 
The obtained function $\sum_{u \in \mathcal{D}} w(u) \phi_{u}$ is called the persistence surface of the persistence diagram $\mathcal{D}$.

\begin{figure}[ht]
    \centering
    \includegraphics[scale=0.37]{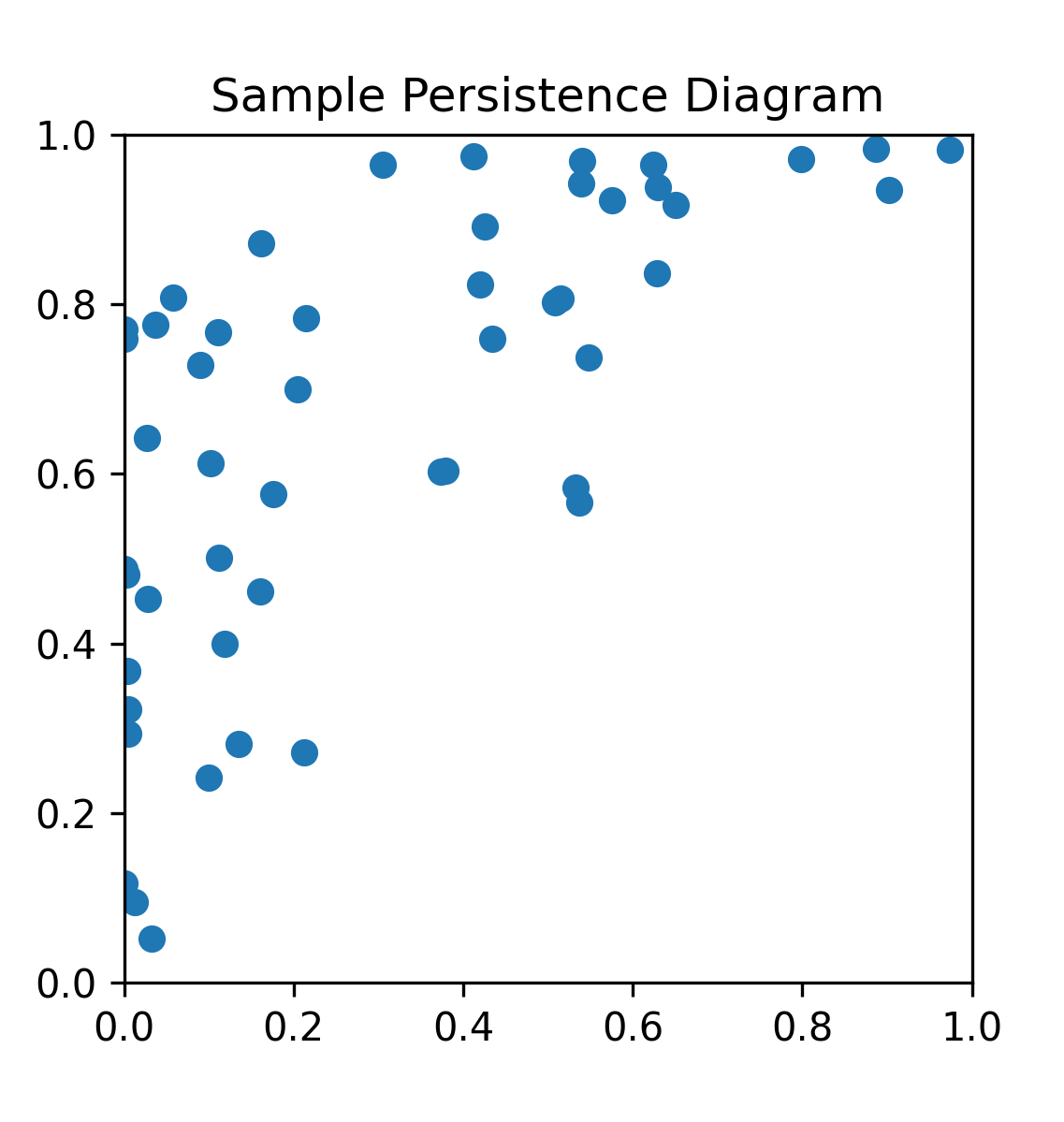}
    \includegraphics[scale=0.37]{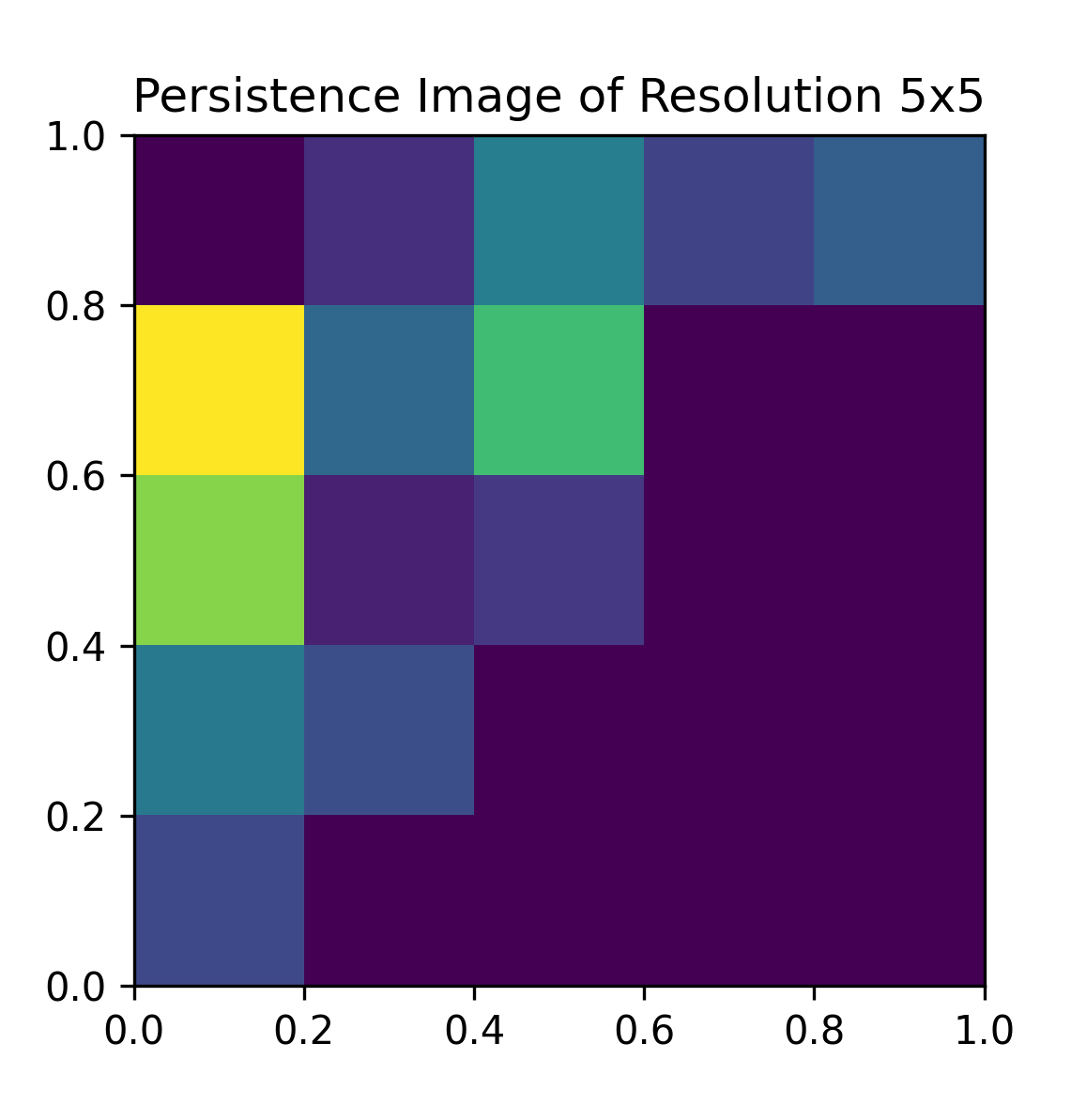}
    \includegraphics[scale=0.37]{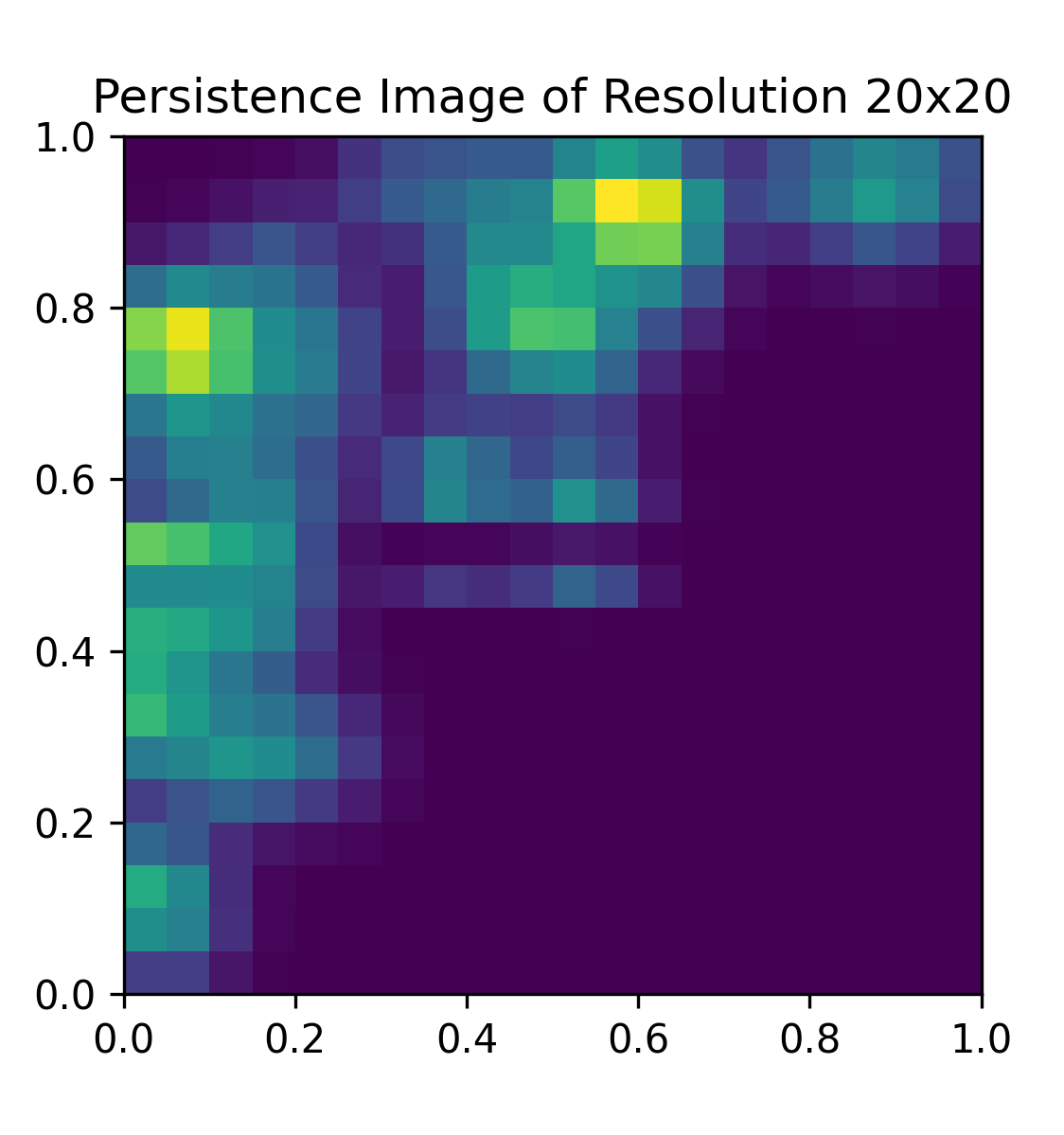}
    \includegraphics[scale=0.37]{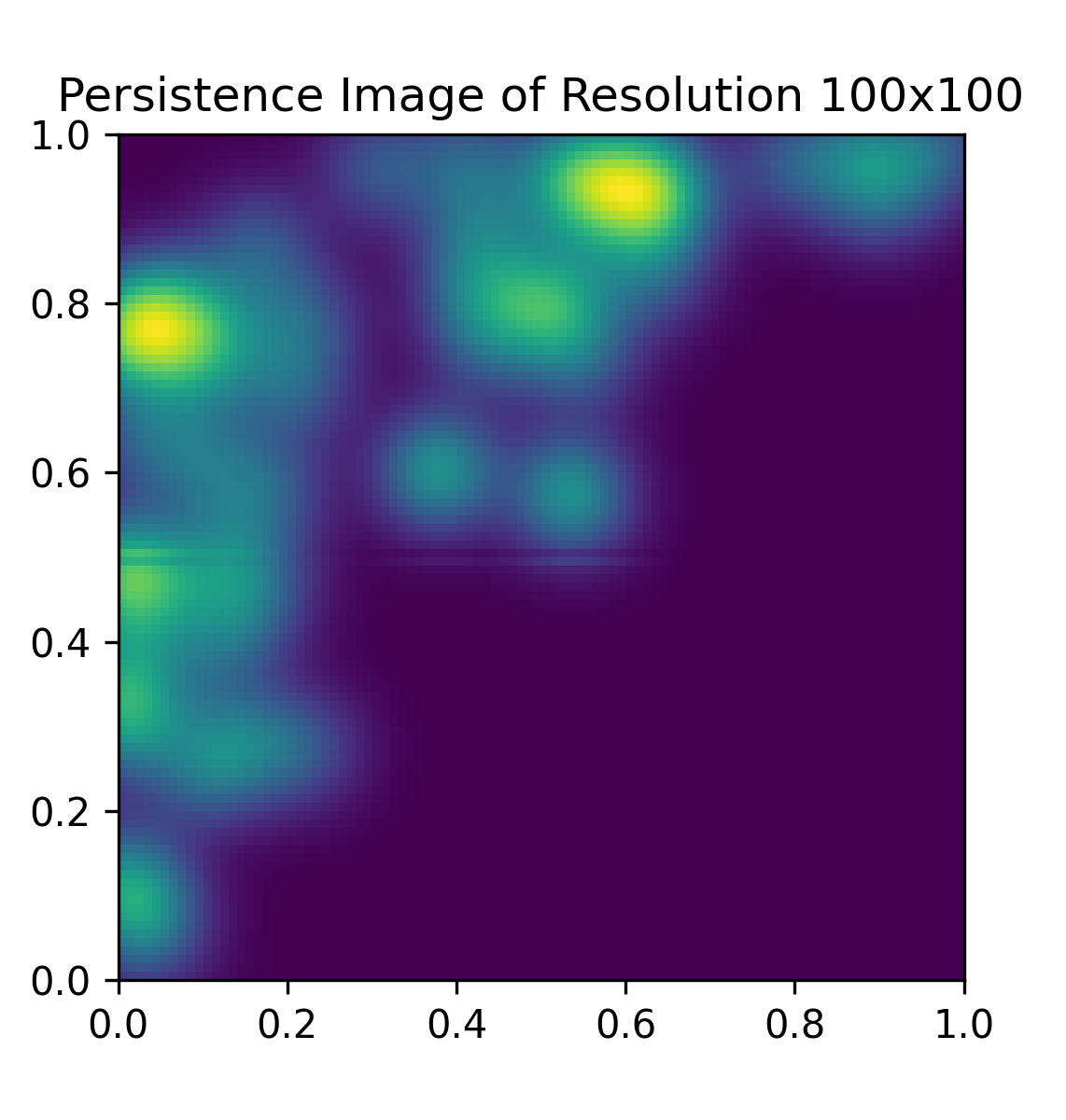}
    \caption{Persistence image of a persistence diagram. 
    From left to right: persistence diagram, persistence images with different resolutions.}
    \label{fig:persistence_image_explanation}
\end{figure}

The last step is to integrate the persistence surface over cells of a grid of dimension $(n_x, n_y)$ with given horizontal and vertical boundaries. 
This grid defines the frame and resolution of the persistence image and has to be chosen (see \cref{fig:persistence_image_explanation}). 
The value of a pixel $R$ of the persistence image of the persistence diagram $\mathcal{D}$ is given by 
\begin{equation*}
    \im(\mathcal{D})(R) = \int_R \sum_{u \in \mathcal{D}} w(u) \phi_{u}(z) dz.
\end{equation*}
The value of the integration over each pixel is stored in a $n_x \times n_y$ image, called the persistence image of the persistence diagram. 

For our application, we consider various image frames, image resolutions (see \cref{tab:parameters_persistence_images}), and weight functions (see \cref{tab:weight_function_persistence_images}). An illustration for the Ordinary filtration can be found in \cref{fig:transformation_procedure_attention_maps_to_persistence_images}.
The tables are in the appendix with examples for all types of persistence images (see \cref{tab:examples_persistence_images_for_all_type}).

\begin{figure}
    \begin{subfigure}{.3\textwidth}
      \centering
      \includegraphics[width=0.9\linewidth]{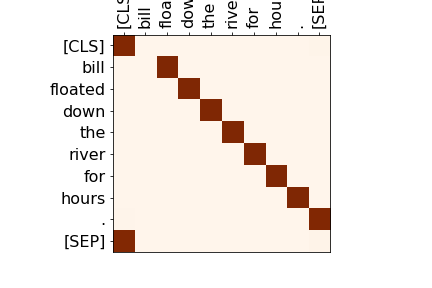}
      \caption{Diagonal attention map \thead{(L: 3, H: 1)}}
      \label{fig:diagonal_attention_map}
    \end{subfigure}
    \hfill
    \begin{subfigure}{.3\textwidth}
      \centering
      \includegraphics[width=.7\linewidth]{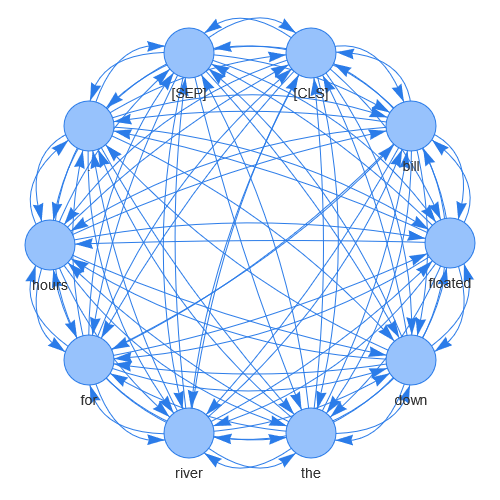}
      \caption{Directed attention graph}
      \label{fig:directed_attention_graph}
    \end{subfigure}
    \hfill
    \begin{subfigure}{.3\textwidth}
      \centering
      \includegraphics[width=0.9\linewidth]{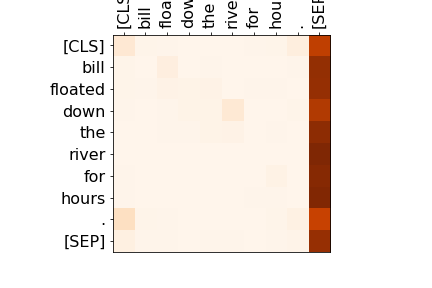}
      \caption{Column attention map \thead{(L: 8, H: 7)}}
      \label{fig:column_attention_map}
    \end{subfigure}

    \begin{subfigure}{\textwidth}
    \begin{center}
        \begin{subfigure}{.3\textwidth}
          \centering
          \includegraphics[width=.7\linewidth]{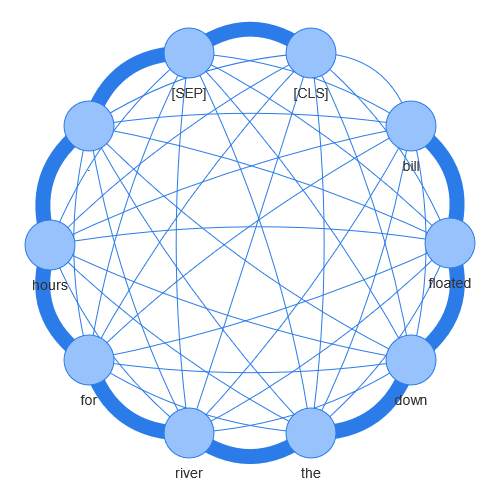}
          \label{fig:diagonal_attention_graph}
        \end{subfigure}
        \hspace{2cm}
        \begin{subfigure}{.3\textwidth}
          \centering
          \includegraphics[width=.7\linewidth]{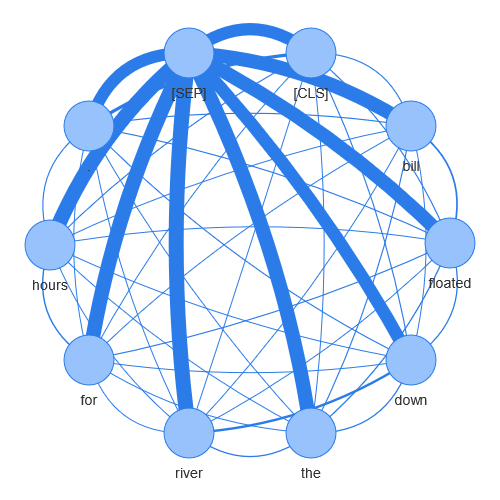}
          \label{fig:column_attention_graph}
        \end{subfigure}
    \caption{Undirected attention graphs where the edge width is proportional to the maximal attention between the two vertices. Left: diagonal (L: 3, H: 1), right: column (L: 8, H: 7).}
    \end{center}
    \label{fig:examples_attention_graphs}
    \end{subfigure}

    \begin{subfigure}{\textwidth}
        \begin{subfigure}{.18\textwidth}
          \centering
          \caption*{$t = 0.0$}
          \includegraphics[width=.8\linewidth]{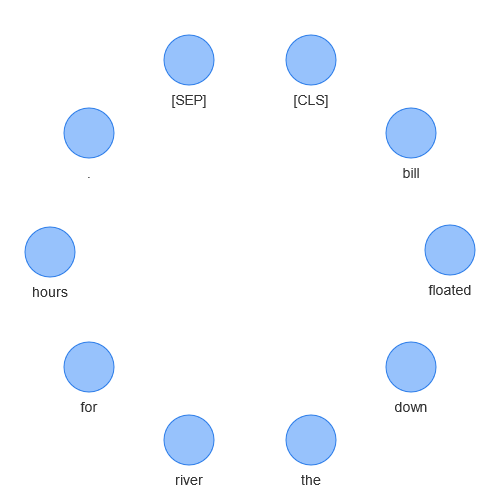}
          \label{fig:diagonal_filtration_0.0}
        \end{subfigure}
        \hfill
        \begin{subfigure}{.18\textwidth}
          \centering
          \caption*{$t = 0.005$}
          \includegraphics[width=.8\linewidth]{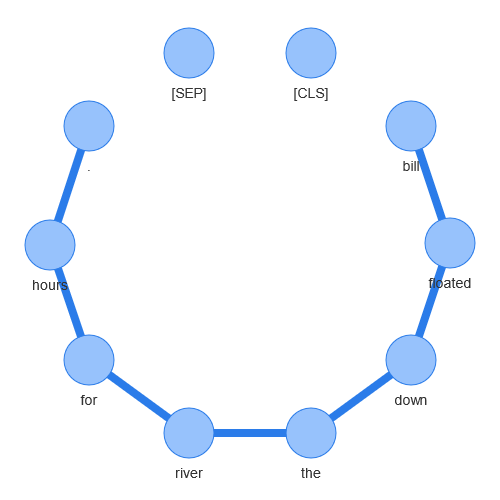}
          \label{fig:diagonal_filtration_0.005}
        \end{subfigure}
        \hfill
        \begin{subfigure}{.18\textwidth}
          \centering
          \caption*{$t = 0.1$}
          \includegraphics[width=.8\linewidth]{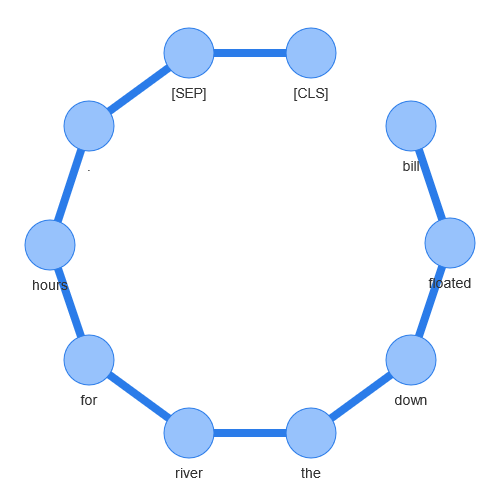}
          \label{fig:diagonal_filtration_0.1}
        \end{subfigure}
        \hfill
        \begin{subfigure}{.18\textwidth}
          \centering
          \caption*{$t = 0.9999$}
          \includegraphics[width=.8\linewidth]{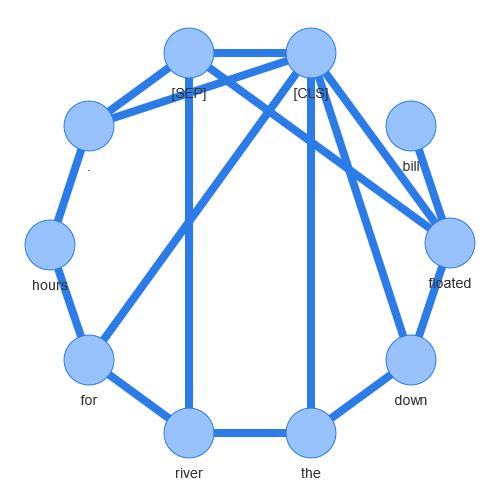}
          \label{fig:diagonal_filtration_0.9999}
        \end{subfigure}
        \hfill
        \begin{subfigure}{.18\textwidth}
          \centering
          \caption*{$t = 1.0$}
          \includegraphics[width=.8\linewidth]{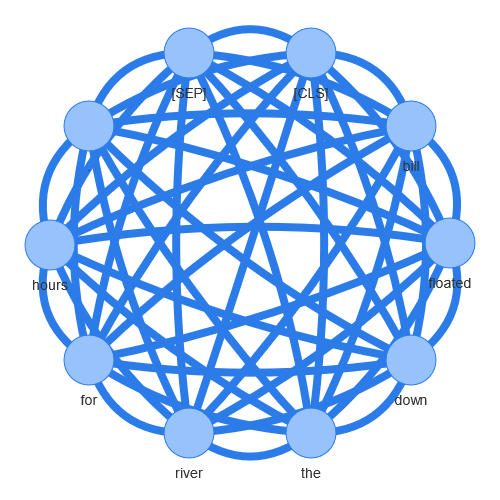}
          \label{fig:diagonal_filtration_1.0}
        \end{subfigure}
    \caption{Attention graph filtration (L: 3, H: 1).}
    \label{fig:diagonal_attention_graph_filtration}
    \end{subfigure}
    
    \begin{subfigure}{\textwidth}
        \begin{subfigure}{.18\textwidth}
          \centering
          \caption*{$t = 0.05$}
          \includegraphics[width=.8\linewidth]{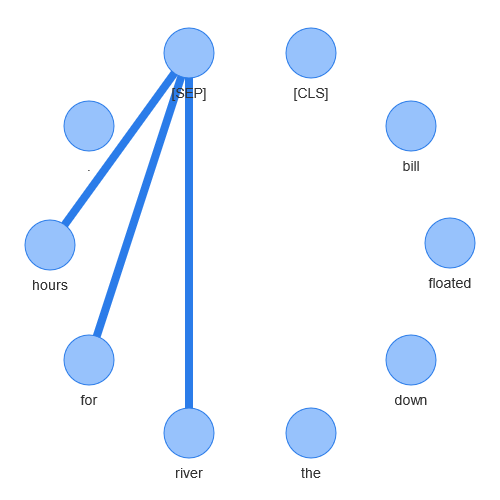}
          \label{fig:column_filtration_0.05}
        \end{subfigure}
        \hfill
        \begin{subfigure}{.18\textwidth}
          \centering
          \caption*{$t = 0.1$}
          \includegraphics[width=.8\linewidth]{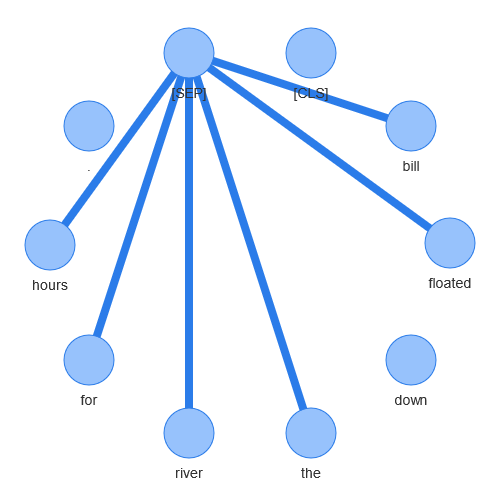}
          \label{fig:column_filtration_0.1}
        \end{subfigure}
        \hfill
        \begin{subfigure}{.18\textwidth}
          \centering
          \caption*{$t = 0.9$}
          \includegraphics[width=.8\linewidth]{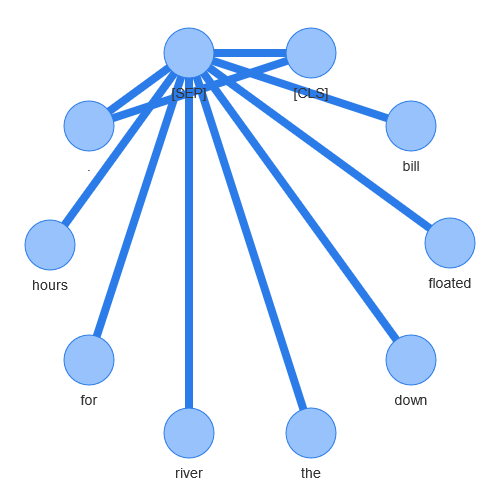}
          \label{fig:column_filtration_0.9}
        \end{subfigure}
        \hfill
        \begin{subfigure}{.18\textwidth}
          \centering
          \caption*{$t = 0.99$}
          \includegraphics[width=.8\linewidth]{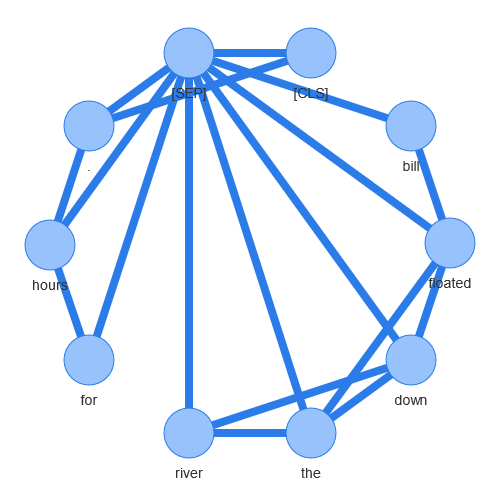}
          \label{fig:column_filtration_0.99}
        \end{subfigure}
        \hfill
        \begin{subfigure}{.18\textwidth}
          \centering
          \caption*{$t = 1.0$}
          \includegraphics[width=.8\linewidth]{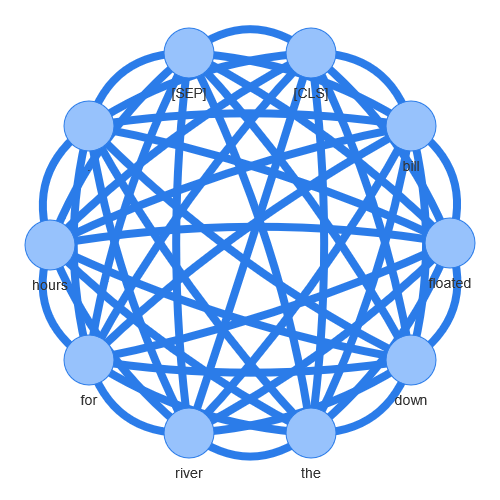}
          \label{fig:column_filtration_1.0}
        \end{subfigure}
    \caption{Attention graph filtration (L: 8, H: 7).}
    \label{fig:column_attention_graph_filtration}
    \end{subfigure}
    
    \begin{subfigure}{.4\textwidth}
        \centering
        \hfill
         \includegraphics[scale=0.35]{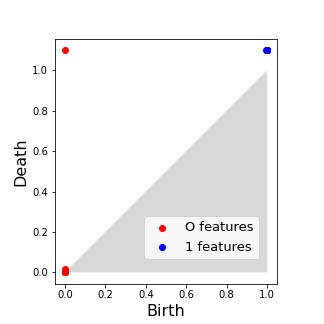}
         \includegraphics[scale=0.16]{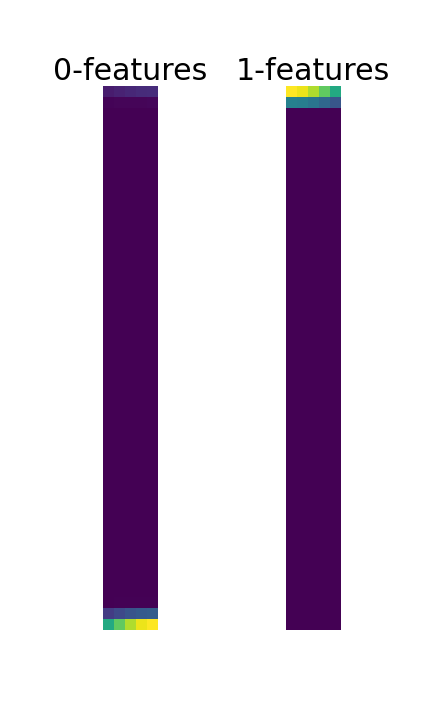}
         \label{fig:diagonal_persistence_diagram_and_image}
         \caption{Persistence diagram (left), persistence images (right), (L: 3, H: 1).}
    \end{subfigure}  
    \hfill
    \begin{subfigure}{.4\textwidth}
        \centering
        \includegraphics[scale=0.35]{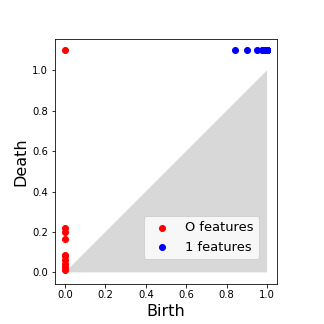}
        \includegraphics[scale=0.16]{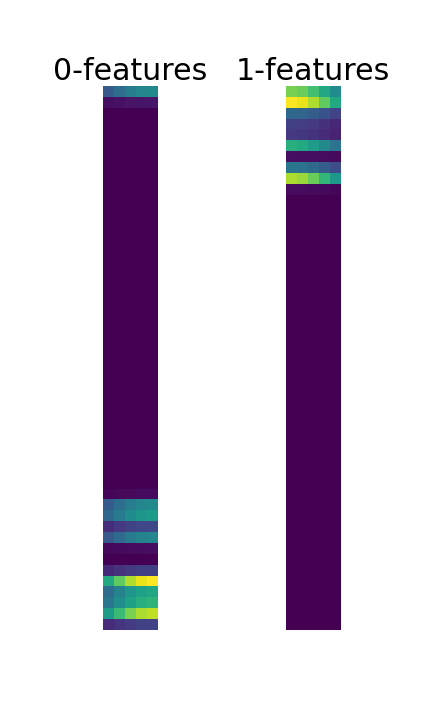}
         \hfill
        \label{fig:column_persistence_diagram_and_images}
        \caption{Persistence diagram (left), persistence images (right), (L: 8, H: 7).}
    \end{subfigure}
    \caption{Transformation from attention maps to persistence images for the sentence \enquote{Bill floated down the river for hours.} and attention heads (L: 3, H: 1) and (L: 8, H: 7). 
    Graph pictures are drawn with from the Pyvis library (\href{https://pyvis.readthedocs.io}{https://pyvis.readthedocs.io}).}
    \label{fig:transformation_procedure_attention_maps_to_persistence_images}
\end{figure}

\section{Methodology}

We compare BERT against the topological model on numerous classification tasks. 
BERT is fine-tuned for a variable amount of epochs. 
To apply the topological model, we first transform each sentence into a stack of persistence images. 
To do so, we feed a fine-tuned BERT model with the sentence and extract the attention graphs for each head. 
We then transform the attention graphs into persistence images. 
One attention graph generates a number of persistence images equals to the number of persistence features inside the filtration (2 for Ordinary, 3 for the two others). 
For the Ordinary filtration, a sentence is transformed into 288 images and for the two others into 432 images.
The topological classifier receives as input a 4-dimensional tensor where the dimensions are:
batch-size, the number of persistence images per sentence, the width and height of the image.

We use the Gudhi library (\cite{gudhi}) to convert persistence diagrams into persistence images for the Ordinary and MultiDim filtrations, and the Giotto-tda library (\cite{giotto-tda}) to manage the directed filtration.
\subsection{Data}
We load the datasets from Hugging Face and we follow the data processing proposed by \cite{betti_bert}. 
All dataset statistics are presented in \cref{Tab:table_data_statistics}.

\begin{description}[wide,itemindent=\labelsep]
\item[CoLA] The Corpus of Linguistic Acceptability (\cite{CoLA}) is part of the GLUE benchmarks (\cite{GLUE}). 
The task is to detect if a sentence is grammatically correct (class 1) or not (class 0). 
We consider the public part of the dataset that contains labels, and disregard the hidden part. 
We use the original validation set for prediction, and we split the train set into train and validation subsets with proportions $90:10$.

\item[IMDB] Large Movie Review Dataset v1.0 (\cite{IMDB}) consists of movie reviews labeled by two sentiments ``positive'' (1) or ``negative'' (0). 
It contains 50,000 reviews. 
We first divide the data set into two equal sized subsets, one for training and the other for testing and validation. 
To obtain attention graphs of manageable sizes, we consider only sentences of length smaller or equal to 128 after tokenization with the standard BERT uncased tokenizer. 
We then divide the second subset into validation and prediction datasets following the proportion $50:50$.

\item[SPAM] The SMS Spam Collection v.1 (\cite{SPAM}) contains text SMS and the task is to determined if they are spam (1) or ham (0). 
It contains 5,574 messages that we divide into train, validation and prediction subsets ($80:10:10$).

\item[SST2] The binary version of Stanford Sentiment Treebank (\cite{SST2}) is also part of the GLUE benchmarks and consists of parts of movie reviews labeled by two sentiment ``positive'' (1) and ``negative'' (0). 
As for the CoLA dataset, we split the original train set into two subsets with proportions $90:10$, and use the original validation set for the prediction.
\end{description}

\begin{table}[ht]
    \centering
        \begin{tabularx}{0.8\textwidth} { 
            >{\raggedright\arraybackslash\hsize=2.0\hsize}X 
            >{\raggedleft\arraybackslash\hsize=.6\hsize}X
            >{\raggedleft\arraybackslash\hsize=.6\hsize}X
            >{\raggedleft\arraybackslash\hsize=.6\hsize}X
            >{\raggedleft\arraybackslash\hsize=.6\hsize}X  }
         \hline
         \hline
          & \textbf{CoLA} & \textbf{IMDB} & \textbf{SPAM} & \textbf{SST2} \\
         \hline
         \textbf{\# Train sent.} & 7695 & 2675 & 4459 & 60614\\
         \textbf{\# Validation sent.} & 856 & 1414 & 557 & 6735\\ 
         \textbf{\# Prediction sent.} & 1043 & 1414 & 558 & 872\\ 
         \textbf{\%} & 70 & 55 & 14 & 51\\
         \textbf{Mean sent. size} & 11 & 88 & 25 & 13\\
         \textbf{Max sent. size} & 47 & 128 & 238 & 64\\
         \hline
         \hline
        \end{tabularx}
    \caption{Statistics of the classification task datasets. \textbf{\%} = percentage of class 1 in prediction set. \textbf{Mean/Max sent. size} = mean/max length of tokenized sentences.}
    \label{Tab:table_data_statistics}
\end{table}

\subsection{Model}
\label{sec}
\begin{figure}[ht]
    \centering
    \includegraphics[scale=0.7]{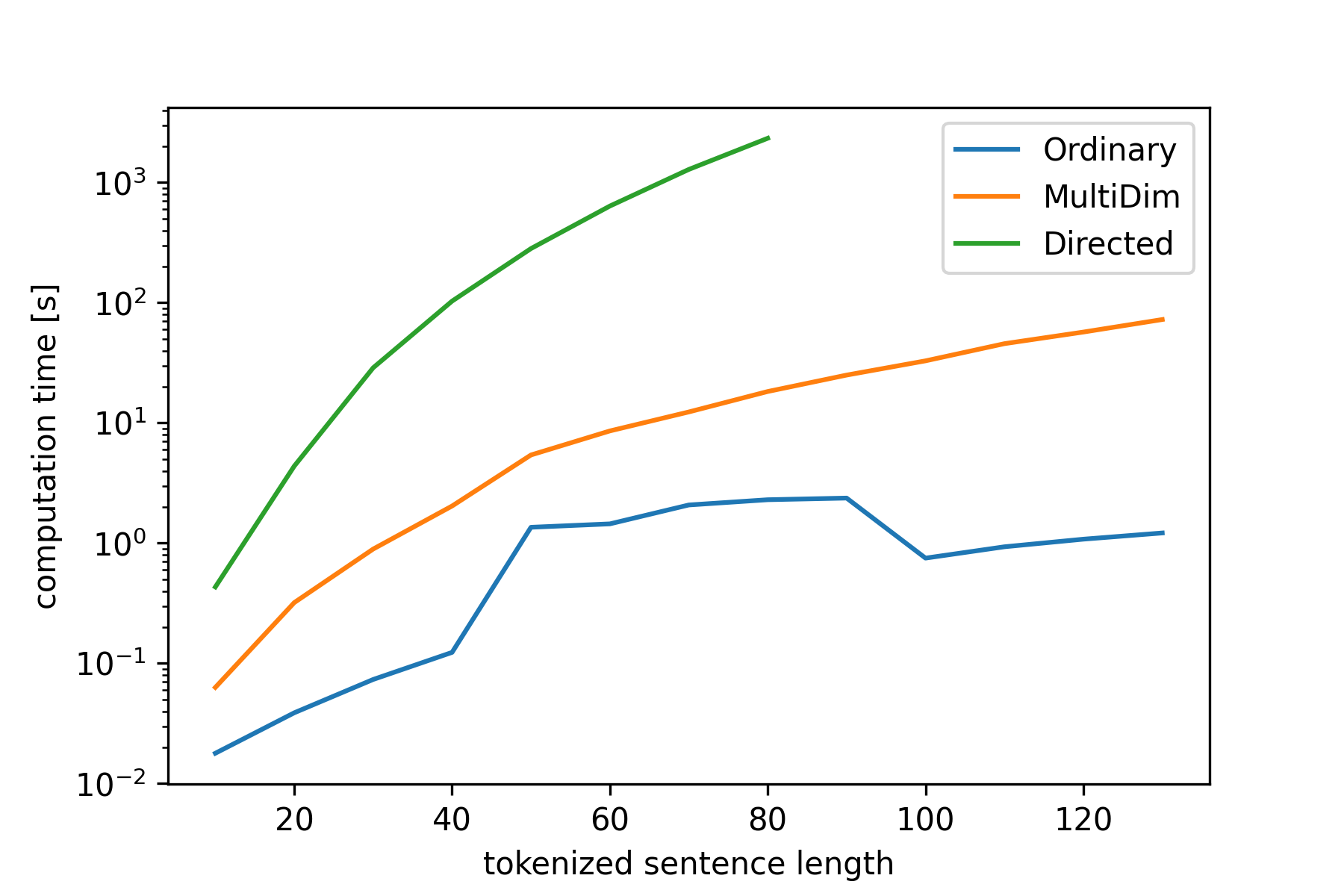}
    \caption{The computation time of the persistence images from the attention maps as a function of the number of tokens in a sequence for the three types of filtrations. Because of the high memory usage for the Directed version, we only plotted sequence lengths of a size up to 80 tokens for it.}
    \label{fig:image_computation_time}
\end{figure}

We run the experiments on the \enquote{bert-base-uncased} model from the HuggingFace library (\cite{huggingface_transformers}) for the BERT baselines. 
For explanatory purposes, we consider numerous variations:

\begin{description}[wide,itemindent=\labelsep]
\item[Fine-tuning] We fine-tuned BERT for 4, 10, and 20 epochs \footnote{We follow the procedure proposed in \href{https://github.com/MohamedAteya/BERT-Fine-Tuning-Sentence-Classification-for-CoLA}{https://github.com/MohamedAteya/BERT-Fine-Tuning-Sentence-Classification-for-CoLA}} 
to study the importance of fine-tuning with respect to the performance of the topological classifiers.
This gives use three BERT models per task, and each is used to transform sentences into persistence images.

\item[Symmetry function] The choice of the symmetry function is a crucial step to get from the attention matrices to the non-directed attention graphs. 
To explore its relevance with respect to the performance of the topological classifier, we use four different symmetry functions: maximum, minimum, multiplication and mean.

\item[Filtration] We also consider the three types of filtrations describe in \cref{sec:persistent_homology}. 
Unfortunately, we are limited in terms of computation time and power as it can reach excessively high values depending on the length of the sentence to transform (see \cref{fig:image_computation_time}). 
Hence for the MultiDim filtration, we consider only the maximum as a symmetry function, and the Directed filtration is only applied to the CoLA dataset.
\end{description}

In total, one fine-tuned BERT model generates 5 or 6 persistence image datasets (PI-datasets). 
We refer to a PI-dataset by the fine-tuned BERT model producing it (name of the dataset and number of fine-tuning epochs), the type of filtration, and the symmetry function, e.g., \enquote{IMDB, 10 epochs, MultiDim, max}. 

Our topological classifiers are convolutional neural networks taking as input the stack of persistence images from one sentence. 
The architecture identified by running a hyperparameter search using the Optuna library (\cite{optuna}) for 500 trials tuning the number of convolution and fully connected layers, the learning rate, the optimizer, and the dropout rate. 
The hyperparameter search is done on one dataset (CoLA, IMDB, SPAM, or SST2), and we say that the model designed from the hyperparameter search result is specific to this dataset. 
We run hyperparameter searches for each classification task and type of filtration. 
The architectures of all the topological classifiers are presented in \cref{tab:architecture_of_topological_models}.

We then investigate the performance of a topological classifier specific to one dataset when evaluated on the other datasets.
To do so, for each PI-dataset, we evaluate the performance of two topological models: 
one whose hyperparameters are optimized on the current dataset, the other one specific to the CoLA dataset. 
We refer to the first model as the specific model and to the second as the general model. 

The inference time of our combination of the BERT model and the topological classifier is greater than the inference time of the BERT model itself.
For example, when using the Ordinary filtration, the time needed to transform a tokenized sentence into a persistence image and predict its class is two times greater than the BERT prediction time.  
It goes up to 20 times for the MultiDim filtration and 70 times for the Directed filtration. The bottleneck is the computation of the persistence images from the persistence diagrams, which is not implemented on GPU but only on CPU.

\subsection{Hardware}
All the computations are done using a virtual machine from the Google Cloud Platform (\href{https://cloud.google.com/}{https://cloud.google.com/}). 
The machine we work on has a  NVIDIA Tesla T4 GPU with 16GB of VRAM, 8 CPUs, named \enquote{n1-standard-8}, with 30GB of RAM.

\section{Results of the topological classifiers}
The performance results obtained by the model whose hyperparameters are optimized on the CoLA dataset and the specific models are outlined in \cref{tab:performance_results} and \cref{tab:performance_results_specific}.
As performance measures, we choose the accuracy on the prediction set and the Matthew Correlation coefficient which is a good metric for unbalanced datasets. 

\begin{landscape}
\begin{table}[p]
    \centering
    \ra{1.3}
    \fontsize{4}{10} \selectfont
    \begin{tabular}{@{}rrrrcrrrcrrrcrrr@{}}
        \toprule
        & \multicolumn{3}{c}{\small\textbf{CoLA}} & \phantom{a} & \multicolumn{3}{c}{\small\textbf{IMDB}} & \phantom{a} & \multicolumn{3}{c}{\small\textbf{SPAM}} & \phantom{a} & \multicolumn{3}{c}{\small\textbf{SST2}}\\
    \cmidrule{2-4} \cmidrule{6-8} \cmidrule{10-12} \cmidrule{14-16}
        & 4 Epochs & 10 Epochs & 20 Epochs && 4 Epochs & 10 Epochs & 20 Epochs && 4 Epochs & 10 Epochs & 20 Epochs && 4 Epochs & 10 Epochs & 20 Epochs\\
    \midrule
    \textbf{BERT} & $0.518~/~80.8$ & $0.561~/~82.2$ & $0.591~/~83.3$ && $\mathbf{0.843~/~92.2}$ & $\mathbf{0.833~/~91.7}$ & $\mathbf{0.831~/~91.7}$ && $\underline{0.993~/~99.8}$ & $\mathbf{1.00~/~100}$  & $\mathbf{1.00~/~100}$ && $\mathbf{0.853~/~92.7}$ & $\mathbf{0.830~/~91.5}$ & $\mathbf{0.830~/~91.5}$\\
    \textbf{Ordinary}\\
    max & $0.539~/~81.2$ & $0.585~/~83.0$ & $0.591~/~83.3$ && $0.776~/~88.8$ & $\underline{0.809~/~90.4}$ & $\underline{0.818~/~91.0}$ && $\mathbf{0.999~/~99.9}$ & \multicolumn{1}{c}{-}  & \multicolumn{1}{c}{-} && $0.799~/~89.8$ & $0.814~/~91.0$ & $0.824~/~91.2$\\
    min & $\underline{0.548~/~81.5}$ & $0.579~/~82.5$ & $0.579~/~82.8$ && $0.591~/~78.4$ & $0.801~/~89.9$ & $0.828~/~90.7$ && $0.980~/~99.5$ & \multicolumn{1}{c}{-}  & \multicolumn{1}{c}{-} && $\underline{0.813~/~90.7}$ & $0.822~/~91.1$ & $\underline{0.825~/~91.2}$\\
    mean & $0.546~/~81.2$ & $\underline{0.586~/~83.0}$ & $\mathbf{0.595~/~83.5}$ && $0.769~/~88.4$ & $0.805~/~90.2$ & $0.813~/~90.7$ && $\underline{0.993~/~99.8}$ & \multicolumn{1}{c}{-}  & \multicolumn{1}{c}{-} && $0.802~/~90.0$ & $0.810~/~90.5$ & $0.805~/~90.2$\\
    mult & $0.524~/~80.8$ & $0.552~/~81.7$ & $0.576~/~82.9$ && $0.130~/~54.2$ & $0.120~/~54.4$ & $0.310~/~64.1$ && $0.978~/~99.5$ & \multicolumn{1}{c}{-}  & \multicolumn{1}{c}{-} && $0.783~/~88.9$ & $0.814~/~90.7$ & $0.820~/~91.2$\\
    \textbf{MultiDim}\\
    max & $\mathbf{0.552~/~81.5}$ & $\mathbf{0.591~/~83.2}$ & $\underline{0.593~/~83.4}$ && $\underline{0.795~/~89.7}$ & $\underline{0.809~/~90.4}$ & $\underline{0.818~/~91.0}$ && $0.978~/~99.5$ & \multicolumn{1}{c}{-}  & \multicolumn{1}{c}{-} && $0.792~/~89.6$ & $\underline{0.824~/~91.2}$ & $0.813~/~90.6$\\
    \textbf{Directed}\\
    - & $0.532~/~80.7$ & $0.577~/~82.6$ & $0.583~/~82.8$ && \multicolumn{1}{c}{-} & \multicolumn{1}{c}{-} & \multicolumn{1}{c}{-} && \multicolumn{1}{c}{-} & \multicolumn{1}{c}{-}  & \multicolumn{1}{c}{-} && \multicolumn{1}{c}{-} & \multicolumn{1}{c}{-} & \multicolumn{1}{c}{-}\\
    \bottomrule
    \end{tabular}
    \caption{Comparison of our classification methods with BERT model (MCC / accuracy on prediction set in \%). 
    The models hyperparameters are optimized on the CoLA dataset.  
    The highest value is in bold, the second highest is underlined. 
    Each value is computed five times. 
    For the datasets CoLA and SST2, the standard deviation of the prediction accuracy of each measure turns around 0.3\%. 
    For the IMDB dataset, the standard deviation turns around 0.8\%. 
    For the SPAM dataset, the standard deviation is below 0.1\%.
    }
    \label{tab:performance_results}
\end{table}

\vspace{20mm}

\begin{table}[p]
    \centering
    \ra{1.3}
    \fontsize{4}{10} \selectfont
    \begin{tabular}{@{}rrrrcrrrcrrrcrrr@{}}
        \toprule
        & \multicolumn{3}{c}{\small\textbf{CoLA}} & \phantom{a} & \multicolumn{3}{c}{\small\textbf{IMDB}} & \phantom{a} & \multicolumn{3}{c}{\small\textbf{SPAM}} & \phantom{a} & \multicolumn{3}{c}{\small\textbf{SST2}}\\
    \cmidrule{2-4} \cmidrule{6-8} \cmidrule{10-12} \cmidrule{14-16}
        & 4 Epochs & 10 Epochs & 20 Epochs && 4 Epochs & 10 Epochs & 20 Epochs && 4 Epochs & 10 Epochs & 20 Epochs && 4 Epochs & 10 Epochs & 20 Epochs\\
    \midrule
    \textbf{BERT} & $0.518~/~80.8$ & $0.561~/~82.2$ & $0.591~/~83.3$ && $\mathbf{0.843~/~92.2}$ & $\mathbf{0.833~/~91.7}$ & $\mathbf{0.831~/~91.7}$ && $\mathbf{0.993~/~99.8}$ & $\mathbf{1.00~/~100}$  & $\mathbf{1.00~/~100}$ && $\mathbf{0.853~/~92.7}$ & $\mathbf{0.830~/~91.5}$ & $\mathbf{0.830~/~91.5}$\\
    \textbf{Ordinary}\\
    max & $0.533~/~80.1$ & $0.585~/~83.0$ & $0.585~/~83.1$ && $0.795~/~89.8$ & $0.793~/~89.6$ & $0.817~/~90.9$ && $0.978~/~99.5$ & \multicolumn{1}{c}{-}  & \multicolumn{1}{c}{-} && $0.793~/~89.5$ & $0.821~/~91.1$ & $0.820~/~91.0$\\
    min & $0.538~/~80.9$ & $0.579~/~82.5$ & $0.579~/~82.8$ && $0.775~/~88.3$ & $\underline{0.818~/~90.9}$ & $\underline{0.810~/~91.6}$ && $0.976~/~99.4$ & \multicolumn{1}{c}{-}  & \multicolumn{1}{c}{-} && $0.785~/~89.2$ & $0.822~/~91.1$ & $\underline{0.824~/~91.1}$\\
    mean & $\underline{0.546~/~81.2}$ & $\underline{0.586~/~83.0}$ & $\mathbf{0.595~/~83.5}$ && $\underline{0.805~/~90.3}$ & $0.805~/~90.2$ & $0.810~/~90.5$ && $\underline{0.987~/~99.7}$ & \multicolumn{1}{c}{-}  & \multicolumn{1}{c}{-} && $\underline{0.804~/~90.2}$ & $\underline{0.823~/~91.1}$ & $0.819~/~90.9$\\
    mult & $0.523~/~80.7$ & $0.552~/~81.7$ & $0.579~/~82.9$ && $0.571~/~77.0$ & $0.552~/~75.6$ & $0.721~/~85.5$ && $0.960~/~99.0$ & \multicolumn{1}{c}{-}  & \multicolumn{1}{c}{-} && $0.799~/~89.9$ & $0.815~/~90.7$ & $0.816~/~90.7$\\
    \textbf{MultiDim}\\
    max & $\mathbf{0.552~/~81.5}$ & $\mathbf{0.591~/~83.2}$ & $\underline{0.593~/~83.4}$ && $0.781~/~88.6$ & $0.788~/~89.4$ & $0.801~/~90.0$ && $0.985~/~99.6$ & \multicolumn{1}{c}{-}  & \multicolumn{1}{c}{-} && $0.793~/~89.6$ & $\mathbf{0.830 / 91.5}$ & $0.813~/~90.6$\\
    \bottomrule
    \end{tabular}
    \caption{Comparison of our classification methods with BERT model (MCC / accuracy on prediction set in \%). 
    The models hyperparameters are optimized on the same dataset task the model is evaluated. 
    The highest value is in bold, the second highest is underlined. 
    Each value is computed five times. 
    For the datasets CoLA and SST2, the standard deviation of the prediction accuracy of each measure turns around 0.3\%. 
    For the IMDB dataset, the standard deviation turns around 0.5\%.
    For the SPAM dataset, the standard deviation is below 0.2\%.
    }
    \label{tab:performance_results_specific}
\end{table}

\end{landscape}

The general topological classifier outperforms BERT for the CoLA dataset by $1\%$ and obtains similar performances for the other datasets. 
It suggests that the persistence images contain as much syntactic information as the encoding provided by BERT. 
The biggest increase in performance is generally obtained by the MultiDim filtration. 
It seems that the more topological information is provided to the model, the better it performs. 
However, the enhancement is not comparable with the computation cost needed to compute the persistent homology of the MultiDim filtration. 
The model based on the Directed filtration performs as well as the Ordinary. 
One explanation could be that the persistence images produced by the Giotto TDA library have different ranges making it difficult for the CNN to compare them.

There is no symmetry function that works best in all cases. 
The mean is the best choice when considering the CoLA dataset. 
The multiplication performs less well for IMDB, but better for the three others. 
In general, all symmetry functions perform similarly on a given task.
Hence this choice is not crucial for the overall performance of the topological classifier.
Interestingly, there is also no significant difference between the results obtained by the specific models versus the general model. 
In some cases, when applied on another dataset, the general model outperforms the specific model. 
This observation would suggest that we could use one topological classifier for multi-task learning with a performance similar to BERT. 

Lastly, there is an overall tendency for performance-boosting when we increase the amount of BERT fine-tuning epochs. 
This is the case on each task with respect to any filtration and any symmetry function, even when the BERT model overfits by training for more epochs (SST2 and IMDB).

\section{UMAP Description}
It is challenging to understand how BERT learns to solve a task, and many attempts have been published (\cite{attention_is_not_explanation, what_does_bert_look_at, bertology}). 
Persistence diagrams of the attention graphs offer a new perspective to look at attention heads.

To do so, we use the UMAP (Uniform Manifold Approximations \& Projection) library (\cite{umap}). It allows to project a high dimensional point cloud onto the two dimensional plane. 
The data we use consists of the persistence diagrams of the 144 attention heads corresponding to one sentence. 
To give a graph structure to the set of data points, we compute the $1$-Wasserstein distance between each pair of diagrams.

\begin{definition}[\cite{computational_topology_for_data_analysis}, Definition 3.9 and 3.10]
 Let $p\geq 1$ be fixed and let $D_1$ and $D_2$ be two finite persistence diagrams of the same homology dimension. Let $D_1'$ and $D_2'$ be the diagrams obtained from $D_1$ and $D_2$ by adding all points on the diagonal with infinite multiplicity.
We define the \textbf{$p$-Wasserstein distance} between $D_1$ and $D_2$ by
  \begin{equation*}
    W_p(D_1,D_2):=\min_{\pi\in \Pi(D_1',D_2')} \Big ( \sum_{x\in D_1'} |x-\pi (x)|^p \Big ) ^{1/p},
  \end{equation*}
  where $\Pi(D_1',D_2')$ is the set of all bijections between $D_1'$ and $D_2'$.

\label{def:wasserstein_distances}
\end{definition}

\begin{figure}[ht]
\begin{subfigure}{.33\textwidth}
  \centering
  \includegraphics[width=.95\linewidth]{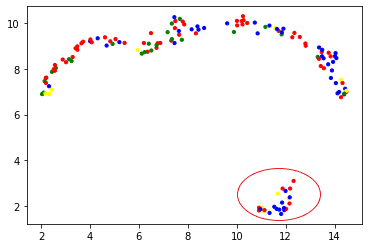}
  \caption{Sentence 1}
  \label{fig:sent_1_cola_ordinary_0}
\end{subfigure}%
\begin{subfigure}{.33\textwidth}
  \centering
  \includegraphics[width=.95\linewidth]{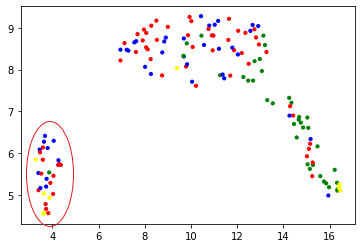}
  \caption{Sentence 2}
  \label{fig:sent_2_cola_ordinary_0}
\end{subfigure}
\begin{subfigure}{.33\textwidth}
  \centering
  \includegraphics[width=.95\linewidth]{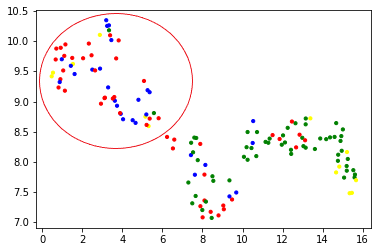}
  \caption{Sentence 3}
  \label{fig:sent_3_cola_ordinary_0}
\end{subfigure}
\caption{UMAP projections for the 0-features of the Ordinary filtration. Clusters are circled in red. They contain 23, 30, 59 elements for sentences 1, 2 and 3 respectively. Cluster 2 shares 20 elements with cluster 1, and cluster 3 contains all of them.}
\label{fig:umap_cola_ordinary_0}
\end{figure}

From the pair-wise distances of the diagrams, UMAP construct a graph as follows: 
It connects each point to a fixed number of closest neighbors. 
Then it projects this graph onto the two-dimensional plane. 
The number of neighbors to connect to has to be chosen manually. 
The higher it is, the more global information will be retrieved. 
The lower it is, the more local information will be displayed in the final projection.
\cref{fig:umap_cola_ordinary_0} shows examples of UMAP projections.

UMAP provides us with information about similitude between persistence diagrams. 
The main observation is the formation of clusters, indicating that there are groups of diagrams that look similar and that are different from the diagrams not in the cluster. 

The primary motivation to use UMAP projections of persistence diagrams is to give another perspective on the self-attention patterns depicted in the paper (\cite{dark_secret_of_bert}). 
We want to see if the persistence diagrams are also clustered in a similar way. 
We consider sentences of the CoLA dataset, and we plot the UMAP for any category of diagrams; one per type of filtration (Ordinary, MultiDim, or Directed) and per homology dimension (0 and 1 for Ordinary, 0,1 and 2 for MultiDim and Directed). 
We manually identified heads with the pattern type of their self-attention maps for each sentence. 
Then we display the UMAP projections, coloring the dots with respect to the pattern: yellow for \textit{diagonal}, red for \textit{vertical}, blue for \textit{diagonal and vertical} and green for \textit{heterogeneous}. 

\begin{figure}[ht]
\begin{subfigure}{.33\textwidth}
  \centering
  \includegraphics[width=.95\linewidth]{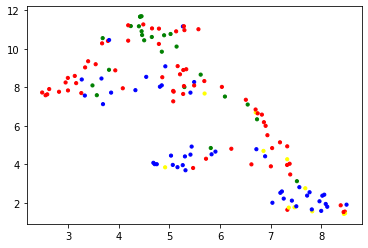}
  \caption{Sentence 1}
  \label{fig:sent_1_cola_ordinary_1}
\end{subfigure}%
\begin{subfigure}{.33\textwidth}
  \centering
  \includegraphics[width=.95\linewidth]{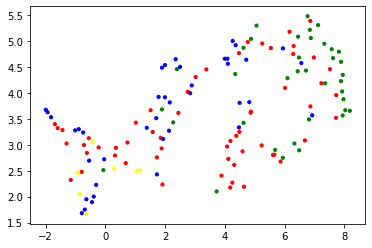}
  \caption{Sentence 2}
  \label{fig:sent_2_cola_ordinary_1}
\end{subfigure}
\begin{subfigure}{.33\textwidth}
  \centering
  \includegraphics[width=.95\linewidth]{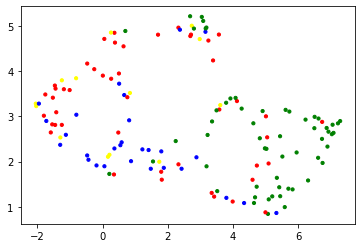}
  \caption{Sentence 3}
  \label{fig:sent_3_cola_ordinary_1}
\end{subfigure}
\caption{UMAPs for the 1-features of the Ordinary filtration.}
\label{fig:umap_cola_ordinary_1}
\end{figure}

In \cref{fig:umap_cola_ordinary_0,,fig:umap_cola_ordinary_1,,fig:umap_cola_twodim_1,,fig:umap_cola_twodim_2,,fig:umap_cola_directed_1,,fig:umap_cola_directed_2}, 
the distribution of the persistence diagrams are not correlated with the distribution of the attention map classes.
The colors are spread evenly across all the points, with no clear monochromatic area. 
In some projection maps, a color gradient between green and red can be observed, but with no clear distinction between the two groups. 
Nevertheless, there are some cluster formations shared across sentences.
In the next subsections we explore the cluster compositions and compare them across sentences. 
We observe similar pattern for a large range of sentences, but only present the plots of three sentences for illustratory purposes. The three sentences are:
\begin{enumerate}
    \item ``Our friends won't buy this analysis, let alone the next one we propose.'' (1)
    \item ``I know a boy mad at John.'' (1)
    \item ``Mary has more friends that two.'' (0)
\end{enumerate}

We only look at the Ordinary filtration here. The UMAP projection for the MultiDim and the Directed filtrations can be found in Appendix \ref{appendix_multi_dim_umap} and Appendix \ref{appendix_directed_umap}.

\cref{fig:umap_cola_ordinary_0} shows the UMAPs of the 0-dimension diagrams for the three considered sentences.

A small cluster is present for the two first sentences.
The overall shape and the color distribution are similar across sentences and we observed that pattern for many more sentences. Looking at the persistent images of each head, no clear pattern is depicted between diagrams in or outside the cluster. 
But surprisingly, the cluster is formed on average by the same heads. For example, the cluster for the second sentence contains 20 of the 23 heads of the cluster from sentence 1.
The circled area for sentence 3 contains all the elements of this cluster of size 23. 

We observe no common cluster for the first homological dimension diagrams, but their UMAP projections tend to divide green and red points.

\ifthenelse{\boolean{is_thesis}}{\section{Discussion}}{\subsection{Discussion}}
The classes proposed by \cite{dark_secret_of_bert} are not observed through the topological lens in these three sentences. 
However, the UMAP projections display clusters that appear for each sentence and whose composition is shared in each example. 
For persistence diagrams of homology dimension zero, there is a small dense cluster shared across sentences, but we could not find a pattern shared between the diagrams inside. 
For homology dimension one, there is a cluster whose diagrams contain mostly short lifespan features, i.e., persistence features with death-time equal or very close to the birth-time. 
And for homology dimension two, there is a cluster whose diagrams present almost only features with a high birth-time.
These specific clusters vary in size, but their composition is shared across sentences.
In average, more than 90\% of the smallest specific cluster is shared with the clusters in the other sentences.
We made the same observation for numerous sentences of the CoLA dataset, leading to the claim that the attention heads of the fine-tuned BERT model specialize in searching for specific information.
This claim was acknowledged in (\cite{bertology}), and we provide a new approach to back it up.

\section{Pruning the heads}\label{sec:pruning_the_images}
Instead of exploring the structure of the persistence diagrams, we investigate what part of the input is most relevant for the model to predict the class of the sentence. 
To do so, we develop a method inspired by GradCam (\cite{gradcam}). 

GradCam is used to help understand the decisions made by deep learning models. 
Given an image, GradCam produces a heatmap that shows how the model makes its decision for that image. 
In our case, as the input is composed of either 288 (twice the number of heads) or 432 images (three times the number of heads), we can not directly apply the method proposed by \cite{gradcam}. 
Instead, we compute the gradient of the output logit with respect to the input image. 
This yields a tensor of the same shape as the input (for example for the ordinary case, the shape is [288, 50, 5]). 
Then we average the absolute value of the gradient over each channel to obtain a number that represents the influence of each individual image on the model output. 
Finally, we take the mean of these values corresponding to images coming from the same attention head (two images in the case of the ordinary filtration, three for the others). 
We end up with a score for the 144 heads of BERT for one input sentence. 
We perform this procedure for a large number of sentences and average all the obtained scores. 

\begin{figure}
    \begin{subfigure}{.3\textwidth}
      \centering
      \includegraphics[width=.95\linewidth]{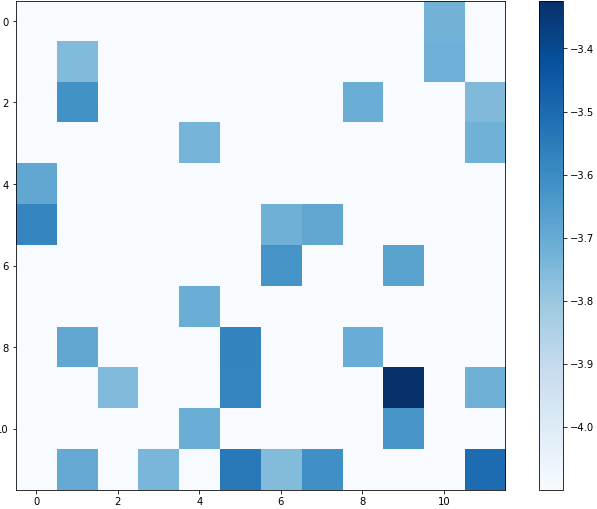}
      \caption{max and 50 sentences}
      \label{fig:30_best_heads_cola_model_4_ordinary_max_50_sent}
    \end{subfigure}
    \hfill
    \begin{subfigure}{.3\textwidth}
      \centering
      \includegraphics[width=.95\linewidth]{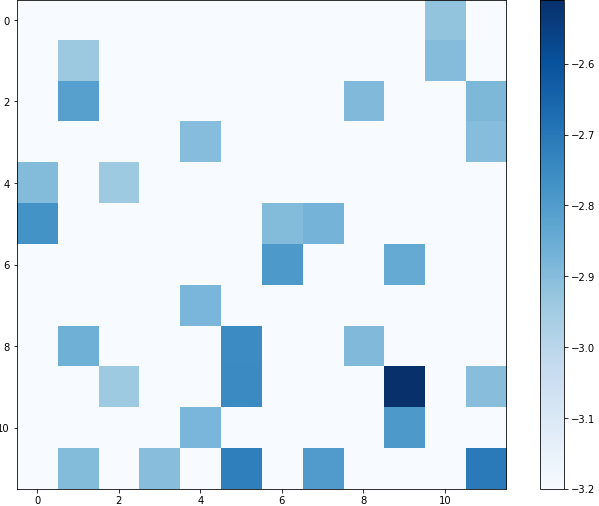}
      \caption{max and 500 sentences}
      \label{fig:30_best_heads_cola_model_4_ordinary_max_500_sent}
    \end{subfigure}
    \hfill
    \begin{subfigure}{.3\textwidth}
      \centering
      \includegraphics[width=.95\linewidth]{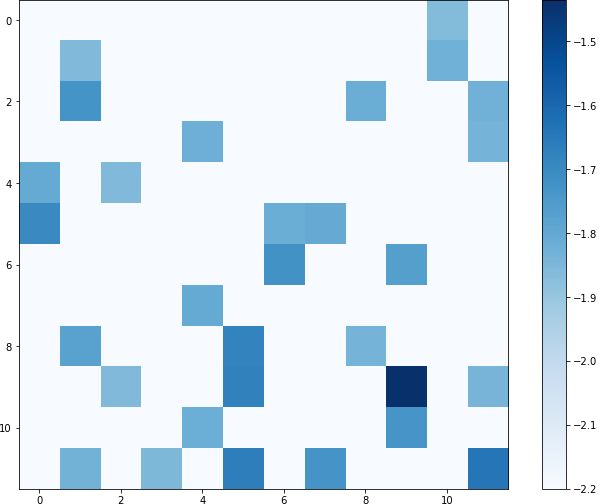}
      \caption{max and 5000 sentences}
      \label{fig:30_best_heads_cola_model_4_ordinary_max_5000_sent}
    \end{subfigure}

    \begin{subfigure}{.3\textwidth}
      \centering
      \includegraphics[width=.95\linewidth]{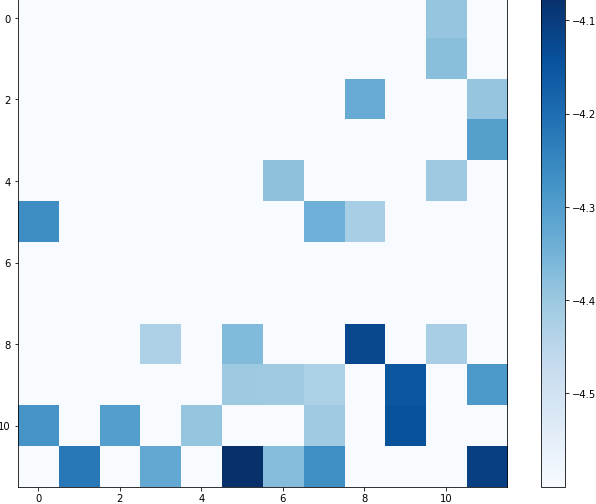}
      \caption{min and 50 sentences}
      \label{fig:30_best_heads_cola_model_4_ordinary_min_50_sent}
    \end{subfigure}
    \hfill
    \begin{subfigure}{.3\textwidth}
      \centering
      \includegraphics[width=.95\linewidth]{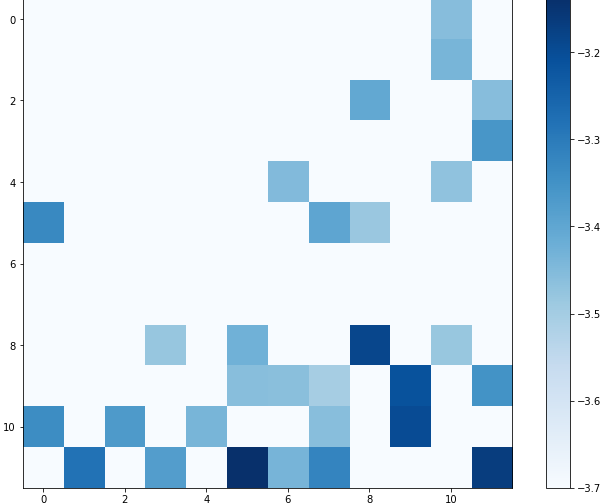}
      \caption{min and 500 sentences}
      \label{fig:30_best_heads_cola_model_4_ordinary_min_500_sent}
    \end{subfigure}
    \hfill
    \begin{subfigure}{.3\textwidth}
      \centering
      \includegraphics[width=.95\linewidth]{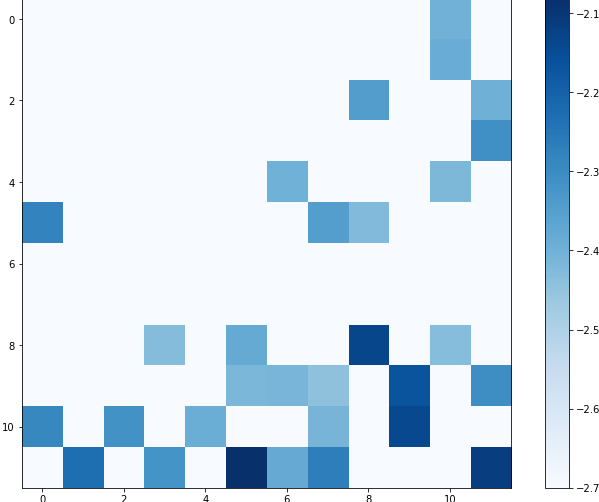}
      \caption{min and 5000 sentences}
      \label{fig:30_best_heads_cola_model_4_ordinary_min_5000_sent}
    \end{subfigure}

    \begin{subfigure}{.3\textwidth}
      \centering
      \includegraphics[width=.95\linewidth]{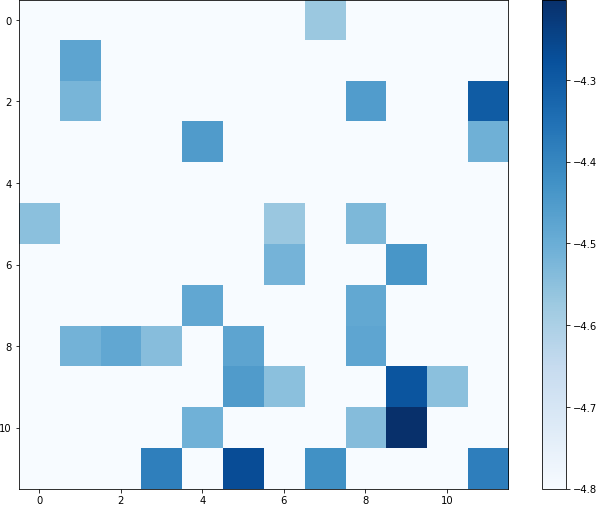}
      \caption{mean and 50 sentences}
      \label{fig:30_best_heads_cola_model_4_ordinary_mean_50_sent}
    \end{subfigure}
    \hfill
    \begin{subfigure}{.3\textwidth}
      \centering
      \includegraphics[width=.95\linewidth]{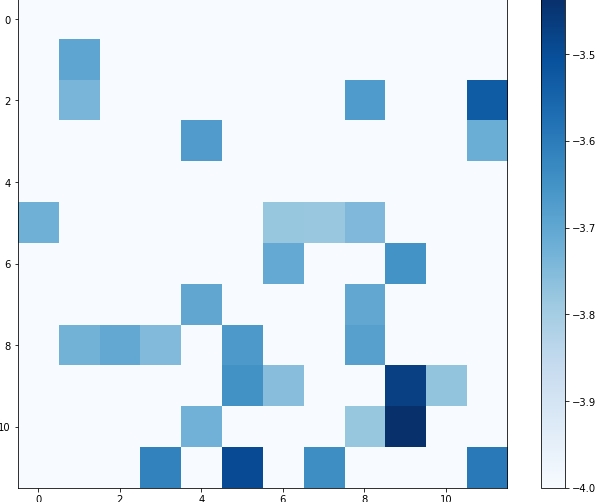}
      \caption{mean and 500 sentences}
      \label{fig:30_best_heads_cola_model_4_ordinary_mean_500_sent}
    \end{subfigure}
    \hfill
    \begin{subfigure}{.3\textwidth}
      \centering
      \includegraphics[width=.95\linewidth]{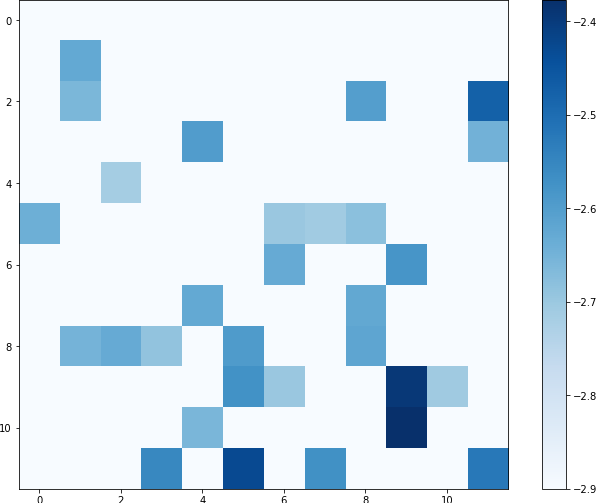}
      \caption{mean and 5000 sentences}
      \label{fig:30_best_heads_cola_model_4_ordinary_mean_5000_sent}
    \end{subfigure}

    \caption{Scores of the 30 best heads for variations of models over the PI-dataset \enquote{CoLA, 4 Epochs, Ordinary, max}. The x-axis is the head id and the y-axis is the layer id.}
    \label{fig:best_heads_scores}
\end{figure}

Our hypothesis is that the higher the score of a head, the more relevant it is to the topological model, and the more information about the sentence structure with respect to the current task it contains. 
\cref{fig:best_heads_scores} displays the 30 best attention heads of the topological model trained on the PI-dataset \enquote{CoLA, 4 Epochs, Ordinary}, but with various numbers of sentences considered and for different symmetry functions.

The heads with the highest scores are independent of the number of sentences used, as there is almost no difference in \cref{fig:30_best_heads_cola_model_4_ordinary_max_50_sent,fig:30_best_heads_cola_model_4_ordinary_max_500_sent,fig:30_best_heads_cola_model_4_ordinary_max_5000_sent}, \cref{fig:30_best_heads_cola_model_4_ordinary_min_50_sent,fig:30_best_heads_cola_model_4_ordinary_min_500_sent,fig:30_best_heads_cola_model_4_ordinary_min_5000_sent}, and \cref{fig:30_best_heads_cola_model_4_ordinary_mean_50_sent,fig:30_best_heads_cola_model_4_ordinary_mean_500_sent,fig:30_best_heads_cola_model_4_ordinary_mean_5000_sent}.
We also observed that the best performing heads are almost independent of the symmetry function considered.
In general, the best heads are located in the deep layers of BERT. 
Therefore, the heads of BERT located in the later layers are the ones that change the most when fine-tuning (\cite{dark_secret_of_bert}), and they are also the most relevant for the topological classifier.

\subsection{Experiments}

We design the following experiment to determine if these high-scoring heads contain most of the necessary information for our topological classifier to perform well.
First we determine the head with the highest scores. 
We train a model on a selected PI-dataset $\alpha$ and we apply our rating procedure on it. 
We look at the $n$ heads with the highest score (for $n$ = 70, 50, 30, 10, 5, 3, 2, and finally the best head).
Then, we train another model on a PI-dataset $\beta$ but with persistence images only related to the $n$ highest scoring heads.
When considering the 70 best heads, we do not consider the persistence images from the other 74 for heads. 
In the case of the Ordinary filtration the input of shape [288, 50, 5] is pruned to the shape [140, 50, 5], as each head produces two persistence images (one for the 0-dimensional features, one for the 1-dimensional features). 
\cref{Tab:table_best_heads_results_base} presents the performance obtained from such a pruning. 
The base PI-dataset considered is \enquote{CoLA, 4 Epochs, Ordinary, max}.
The other columns are variations of the base PI-dataset by changing either the considered symmetry function or the number of fine-tuning epochs. 
In \cref{Tab:table_best_heads_results_base} the PI-dataset $\alpha$ is the base PI-dataset and the PI-dataset $\beta$ is the PI-dataset identified by the column.
In \cref{Tab:table_best_heads_results_base_specific}, $\alpha$ and $\beta$ are both the PI-dataset identified by the column. 

\begin{table}
    \centering
    \begin{tabularx}{1.0\textwidth} { |
            >{\raggedright\arraybackslash}X :
            >{\raggedleft\arraybackslash}X :
            >{\raggedleft\arraybackslash}X :
            >{\raggedleft\arraybackslash}X :
            >{\raggedleft\arraybackslash}X :
            >{\raggedleft\arraybackslash}X |}
         \hline
         \hline
          & \textbf{\thead{CoLA \\ 4 Epochs \\ Ordinary \\ Max}} & \textbf{\thead{CoLA \\ 4 Epochs \\ Ordinary \\ Min}} & \textbf{\thead{CoLA \\ 4 Epochs \\ Ordinary \\ Mean}} & \textbf{\thead{CoLA \\ 10 Epochs \\ Ordinary \\ Max}} & \textbf{\thead{CoLA \\ 20 Epochs \\ Ordinary \\ Max}}\\
          \hline
          \textbf{BERT} & $0.518~/~80.6$ & $0.518~/~80.6$ & $0.518~/~80.6$ & $0.561~/~82.2$ & $\mathbf{0.591~/~83.3}$\\
          \hline
          144 Heads & $0.539~/~81.2$ & $\underline{0.548~/~81.5}$ & $0.546~/~81.2$ & $0.585~/~83.0$ & $\mathbf{0.591~/~83.3}$\\
          70 Heads & $0.557~/~81.3$ & $\mathbf{0.552~/~81.7}$ & $0.544~/~80.6$ & $0.581~/~82.9$ & $0.587~/~83.2$\\   
          50 Heads & $0.557~/~81.2$ & $0.545~/~81.5$ & $0.551~/~81.2$ & $0.580~/~82.7$ & $\underline{0.590~/~83.3}$\\   
          30 Heads & $\mathbf{0.582~/~82.4}$ & $\underline{0.548~/~81.5}$ & $0.548~/~80.9$ & $0.584~/~83.0$ & $0.578~/~82.8$\\
          10 Heads & $0.545~/~81.0$ & $0.540~/~81.2$ & $0.555~/~81.6$ & $0.589~/~83.1$ & $0.574~/~82.5$\\
          5 Heads & $0.553~/~81.7$ & $0.519~/~80.6$ & $\mathbf{0.563~/~81.9}$ & $\underline{0.594~/~83.3}$ & $0.590~/~83.2$\\
          3 Heads & $0.554~/~81.9$ & $0.504~/~80.0$ & $\underline{0.556~/~81.6}$ & $\mathbf{0.594~/~83.4}$ & $0.583~/~82.9$\\
          2 Heads & $\underline{0.560~/~82.1}$ & $0.497~/~79.8$ & $0.525~/~80.3$ & $0.582~/~82.9$ & $0.582~/~82.9$\\
          1 Head & $0.492~/~79.4$ & $0.469~/~78.7$ & $0.479~/~78.7$ & $0.459~/~77.4$ & $0.535~/~80.7$\\
          \hline
          \hline
    \end{tabularx}
    \caption{Performances of the topological models while considering different number of high scoring heads. \textbf{These are determined from the base PI-dataset \enquote{CoLA, 4 Epochs, ordinary, max}.}}
    \label{Tab:table_best_heads_results_base}
\end{table}

\begin{table}
    \centering
    \begin{tabularx}{1.0\textwidth} { |
            >{\raggedright\arraybackslash}X :
            >{\raggedleft\arraybackslash}X :
            >{\raggedleft\arraybackslash}X :
            >{\raggedleft\arraybackslash}X :
            >{\raggedleft\arraybackslash}X :
            >{\raggedleft\arraybackslash}X |}
         \hline
         \hline
          & \textbf{\thead{CoLA \\ 4 Epochs \\ Ordinary \\ Max}} & \textbf{\thead{CoLA \\ 4 Epochs \\ Ordinary \\ Min}} & \textbf{\thead{CoLA \\ 4 Epochs \\ Ordinary \\ Mean}} & \textbf{\thead{CoLA \\ 10 Epochs \\ Ordinary \\ Max}} & \textbf{\thead{CoLA \\ 20 Epochs \\ Ordinary \\ Max}}\\
          \hline
          \textbf{BERT} & $0.518~/~80.6$ & $0.518~/~80.6$ & $0.518~/~80.6$ & $0.561~/~82.2$ & $0.591~/~83.3$\\
          \hline
          144 Heads & $0.539~/~81.2$ & $\underline{0.548~/~81.5}$ & $0.546~/~81.2$ & $0.585~/~83.0$ & $0.591~/~83.3$\\
          70 Heads & $0.557~/~81.3$ & $0.539~/~81.0$ & $0.553~/~81.6$ & $0.571~/~82.6$ & $\mathbf{0.601~/~83.7}$\\   
          50 Heads & $0.557~/~81.2$ & $\mathbf{0.557~/~81.9}$ & $0.556~/~81.7$ & $0.582~/~82.9$ & $0.591~/~83.3$\\   
          30 Heads & $\mathbf{0.582~/~82.4}$ & $0.541~/~81.4$ & $0.537~/~81.2$ & $0.576~/~82.6$ & $0.581~/~82.9$\\
          10 Heads & $0.545~/~81.0$ & $0.547~/~81.4$ & $\underline{0.556~/~81.8}$ & $\underline{0.592~/~83.1}$ & $\underline{0.581~/~83.4}$\\
          5 Heads & $0.553~/~81.7$ & $0.518~/~80.5$ & $0.555~/~81.6$ & $0.587~/~83.0$ & $0.591~/~83.3$\\
          3 Heads & $0.554~/~81.9$ & $0.500~/~79.9$ & $\mathbf{0.569~/~82.1}$ & $\mathbf{0.590~/~83.2}$ & $0.579~/~82.7$\\
          2 Heads & $\underline{0.560~/~82.1}$ & $0.500~/~80.0$ & $0.527~/~80.4$ & $0.578~/~82.7$ & $0.579~/~82.7$\\
          1 Head & $0.492~/~79.4$ & $0.463~/~78.5$ & $0.487~/~79.0$ & $0.458~/~77.8$ & $0.539~/~80.9$\\
          \hline
          \hline
    \end{tabularx}
    \caption{Performances of the topological models while considering different numbers of high scoring heads. \textbf{These are determined specifically for each PI-dataset.}}
    \label{Tab:table_best_heads_results_base_specific}
\end{table}

The performance of the topological classifier trained on the base PI-dataset does not decrease with decreasing number of input images. 
It is even increasing and outperforms the 144 heads model at most by 2\% in accuracy on the prediction set. 
Astonishingly, with 2 heads, our topological classifier outperforms BERT by 1.5\% in accuracy. 
The model receives only 4 images of the initial 288, and its accuracy is still very large. 
It diminishes when considering only one head, but it still has a high accuracy of almost 80\%.

The same trend can be observed for all the other PI-datasets: an increasing or constant accuracy when we decrease the number of considered heads from 70 to 10. However, we get a lower accuracy when we only consider less than ten heads.
In all the models, decreasing the number of input images increases the performance of our topological classifier, up to a certain minimal number of heads considered. 
Choosing a model-specific rating of heads does not change the behavior in the results, neither in the trend nor in values. 
Hence the heads containing the most relevant information with respect to our topological classifier are consistent across different symmetry functions and number of fine-tuning epochs.

We do not observe the same phenomenon when we look at different datasets:
there are not high scoring heads shared across tasks (see \cref{Tab:table_best_heads_results_tasks,,Tab:table_best_heads_results_tasks_specific} in Appendix). 
For the pruning to be efficient, the head scores has to be determined specifically for each dataset.

We also consider the effect of image pruning for other filtrations (see \cref{tab:table_best_heads_results_filtrations} in Appendix) and observe that they also benefit a gain in performance from it. 
Interestingly, to increase the performance of the topological classifier, one should prefer to remove some well-chosen input images, rather than considering more complex and computation-demanding filtrations.

The highest-rated heads may not be the only ones from which the model retrieves valuable information. 
To explore this, we trained the model while keeping the images coming from all the attention heads except the heads with the highest rating (see \cref{Tab:table_best_heads_results_inverted} in Appendix). 
We conclude that the high scoring heads are not necessary. 
Even without them, the topological classifier obtains a similar performance as in the non-pruning case. 

These results verify the statements proposed by \cite{bert_plays_lottery} that one can find a good sub-model inside BERT even when it is highly pruned. 
In \cite{are_sixteen_heads_better_than_one}, some layers where pruned to one head with no effect on performance. 
With our procedure, we could reduce the number of head to five without a loss of performance.

We further investigate how the our pruning procedure behaves across different fine-tuned BERT models in Appendix \ref{appendix_pruning}.

\subsection{Discussion}
From all these experiments, a clear observation arises: specific heads are highly relevant for our model to perform comparably to BERT or even outperform it. 
To investigate what these heads look like, we plot the attention maps (\cite{dark_secret_of_bert}) for the three best heads of our base PI-dataset (\enquote{CoLA, 4 Epochs, Ordinary, max}) for ten sentences in class 1 (grammatically correct) and 10 sentences in class 0 (grammatically incorrect). 

\begin{figure}
    \begin{subfigure}{.95\textwidth}
      \centering
      \includegraphics[width=.95\linewidth]{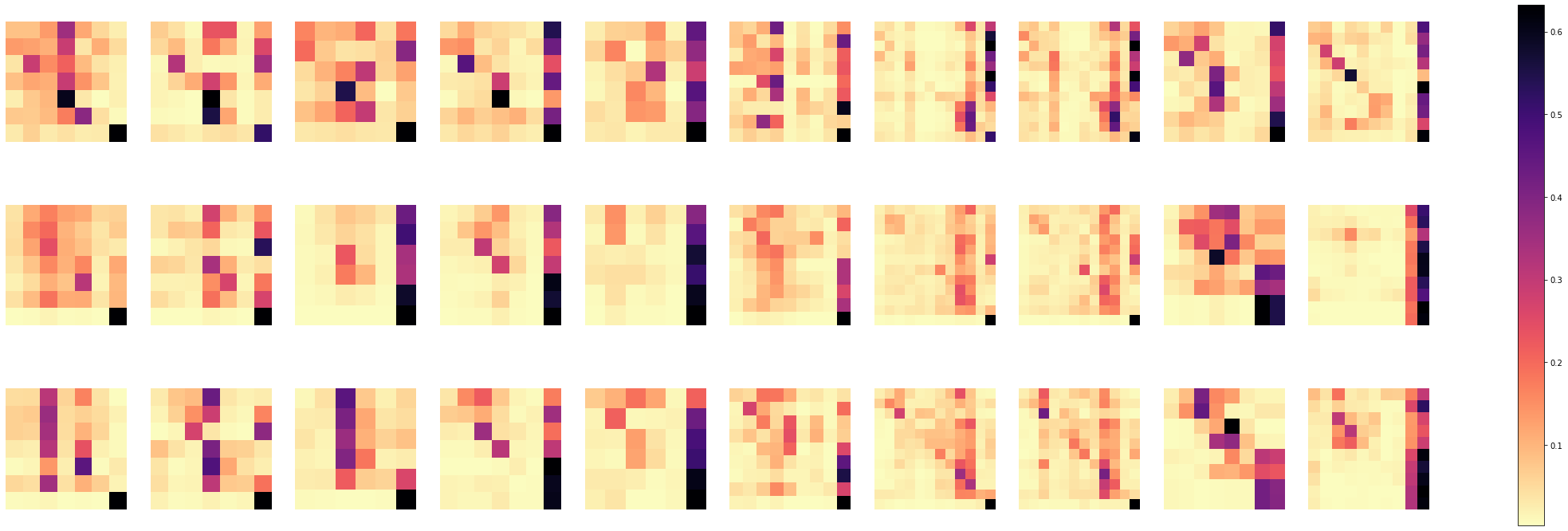}
      \caption{Attention maps corresponding to sentences in class 1.}
      \label{fig:attention_maps_best_3_heads_class_1}
    \end{subfigure}

    \begin{subfigure}{.95\textwidth}
      \centering
      \includegraphics[width=.95\linewidth]{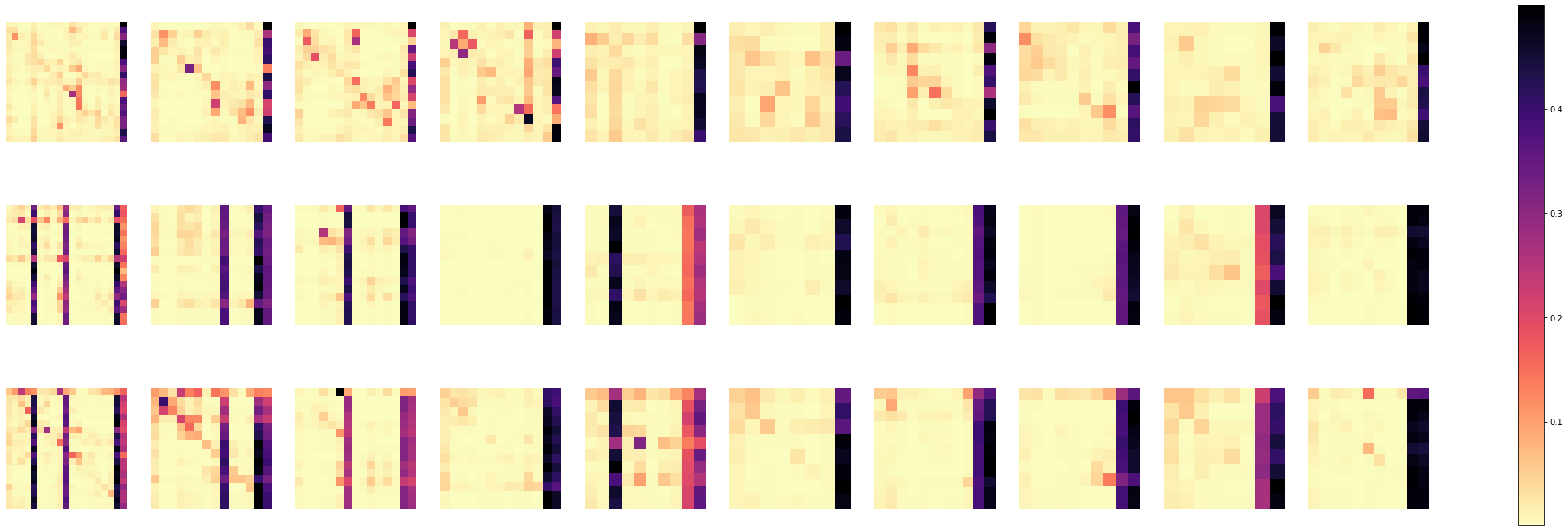}
      \caption{Attention maps corresponding to sentences in class 0.}
      \label{fig:attention_maps_best_3_heads_class_0}
    \end{subfigure}

    \caption{Attention maps of the 3 best heads of the PI-dataset \enquote{CoLA, 4 Epochs, Ordinary, max}. Each line corresponds to a head, the best one at the top. Sentences are part of the CoLA dataset.}
    \label{fig:attention_maps_best_3_heads}
\end{figure}

Almost all the feature maps have high values on the column corresponding to the [SEP] token. 
This means that to encode the sentence, the head will mainly consider the current vectorization of [SEP] to compute the new representation of each token. 
This suggests that this pattern contains sufficient information to get a good performance from the topological classifier. 
In \cite{what_does_bert_look_at}, this peculiar pattern on [SEP] is interpreted as a \textit{no-op} function; the default mode a head enters if it cannot apply its specific function. 
For example in (\cite{what_does_bert_look_at}) the authors found a head that is specialized in  verb-subject recognition. This head puts all the attention on [SEP] if the input word is not a verb. 
Our observation suggests that giving attention to [SEP] contains useful information, and is not only a way for the head to do no operation. 

\begin{figure}
    \centering
    \includegraphics[width=.95\linewidth]{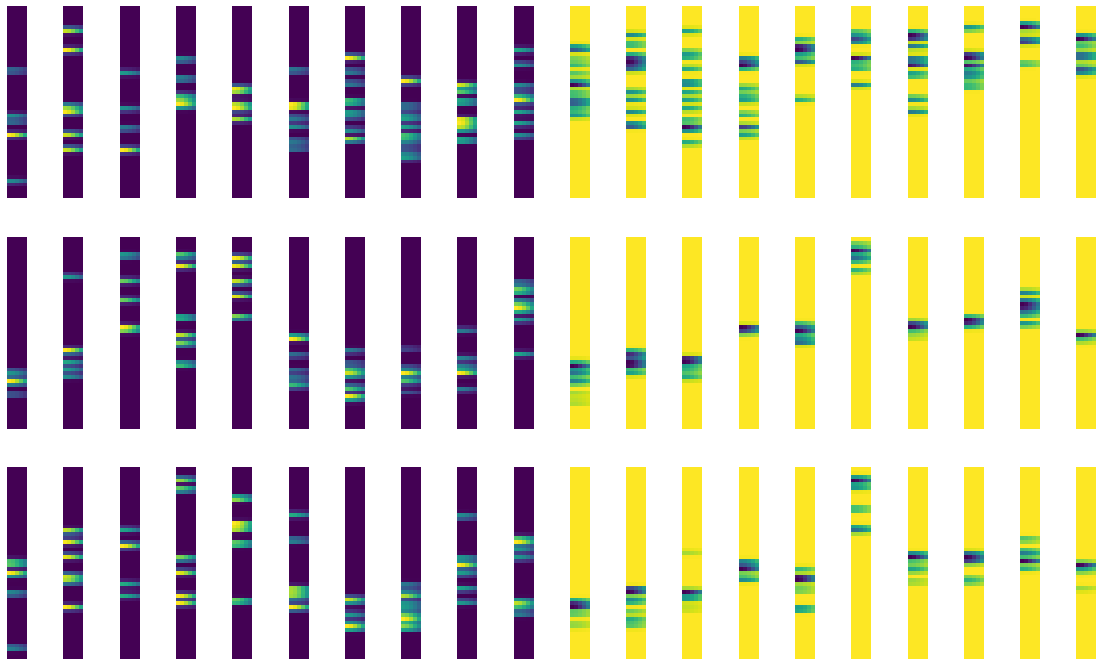}
    \caption{Persistence images of the 3 best heads from the PI-dataset \enquote{CoLA, 4 Epochs, Ordinary, max}. 
    Each line corresponds to a head, the best one at the top. 
    The first 10 pictures refers to sentences in class 0, the last ten to class 1. 
    The persistence images represent the zero dimensional persistence features.}
    \label{fig:persistence_images_best_3_heads}
\end{figure}

To deduce how full attention on [SEP] can be used to classify sentences, the persistence images are of great help. 
\cref{fig:persistence_images_best_3_heads} plots the persistence images corresponding to the above attention maps.

These images correspond to attention almost exclusively on the token [SEP].
Each bar in the persistence image represents the filtration value where a token is connected to the [SEP] vertex. 
Sparse diagrams represent a filtration where the vertices are connected to the [SEP] vertex at different filtration values. Pack diagrams indicate a filtration where the vertices are connected to [SEP] in a narrow range of filtration values. 
The valuable information of the full-attention-to-[SEP] pattern might be in the connection to the [SEP] vertex in the attention graph, which is easily obtained by the topological classifiers.

\begin{figure}
    \centering
    \includegraphics[width=.95\linewidth]{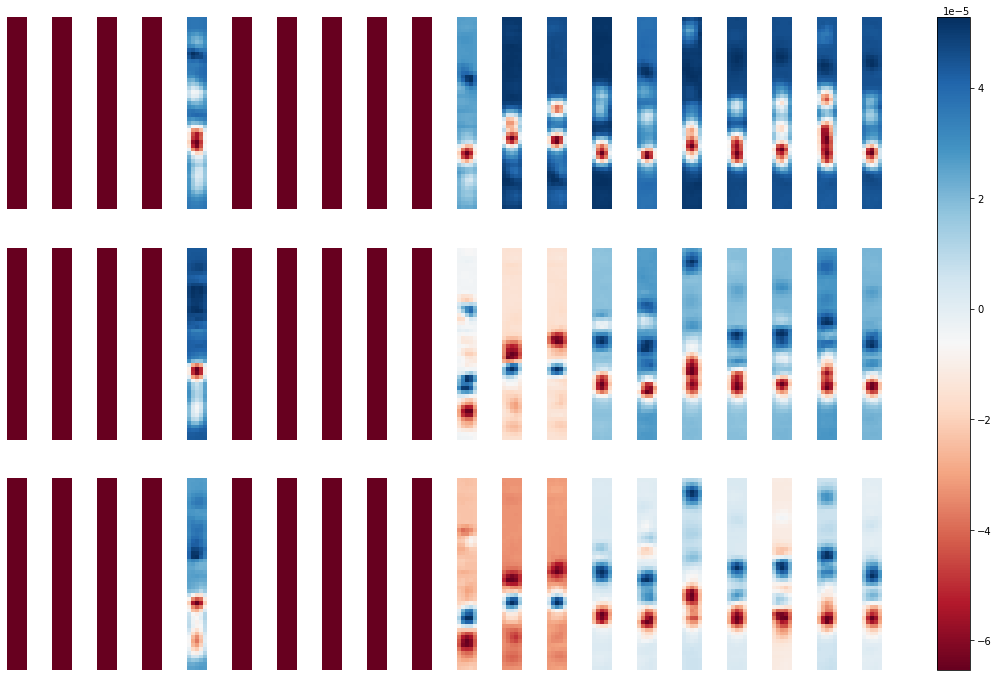}
    \caption{Gradient values of each pixel from the PI-dataset \enquote{CoLA, 4 Epochs, Ordinary, max}.
    Each line corresponds to one head, and the lines are sorted by average importance score. 
    The first 10 pictures refers to sentences in class 0, the last ten to class 1. 
    The darker the red, the more the area influences the model towards class 0; the darker the blue, the more the influence is towards class 1. 
    The persistence images represent the zero dimensional persistence features.}
    \label{fig:gradient_values_best_3_heads}
\end{figure}

But where does the model look when it processes persistence images? 
The regions that are the most relevant to the model appear when the gradient is visualized, as in \cref{fig:gradient_values_best_3_heads}.
The darker the red, the more the area influences the model towards class 0; the darker the blue, the more the influence is towards class 1. 
The white areas are not considered by the model as being relevant to the output classification.. 

The model has different behavior for each sentence label. 
If the sentence is in class 0, any positive pixel value will decrease the model output, thus increasing the probability of class 0. 
This is independent of the death time of the feature, and hence only the existence is relevant. 
If the sentence belongs to class 1, then again, any point in the image will influence the output probability towards class 1, except for some particular regions generally situated at filtration values between $0.7$ and $0.8$, where the influence is inverted. 
Persistence features of dimension 0 that die at this filtration value influence the model output toward class 0. 
A 0-dimensional feature dies when it gets connected to the main connected component; hence if a vertex has edges with the lowest value in this specific filtration range, it will influence the model toward class 0. 

From the CoLA dataset perspective, a sentence containing such a token has a higher chance of being predicted as grammatically incorrect.

\section{Adversarial Attacks}
The transformation of attention head activations into persistence images, despite being computationally demanding could increases the robustness of our model.
On the other hand, pruning the number of heads considered for the input of the model might diminish the stability of the classifier. 
To explore both concerns we face our topological model with adversarial attacks. 
We used TextAttack (\cite{textattack}) to generate hundred attacks for the SST2 dataset.
After removing the skipped and failed attempts, 89 successful attacks on BERT remain.
We then apply the topological classifier \enquote{SST2, 4 Epochs, ordinary, max} on each sentence before and after the changes made by the attacks, and with various numbers of heads considered, determined by the pruning method presented in \cref{sec:pruning_the_images}. 

We consider SST2 and not the CoLA dataset, because the attacks generated by TextAttack were mostly transforming a grammatically correct sentence into a grammatically incorrect one, and the attack is considered a success even if the model detected the grammatical mistake. For that we looked at the 89 sentences in SST2 that where initially correctly classified by the BERT model.

\begin{table}
    \centering
        \begin{tabularx}{0.8\textwidth} { |
            >{\raggedright\arraybackslash\hsize=2\hsize}X |
            >{\raggedleft\arraybackslash\hsize=1.1\hsize}X
            >{\centering\arraybackslash\hsize=0.7\hsize}X |
            >{\raggedleft\arraybackslash\hsize=1.1\hsize}X
            >{\centering\arraybackslash\hsize=0.7\hsize}X
            >{\raggedleft\arraybackslash\hsize=0.4\hsize}X |}
         \hline
         \hline

          & \multicolumn{2}{c}{Avoided Attacks} \vline & \multicolumn{2}{c}{Avoided Common Attacks} & \# \\
         \hline
         144 heads & 46 & (\textit{52\%}) & 40 & (\textit{45\%}) & 83\\
         70 heads & 45 & (\textit{50\%}) & 40 & (\textit{48\%}) & 84\\ 
         50 heads & 42 & (\textit{47\%}) & 36 & (\textit{43\%}) & 83\\ 
         30 heads & 39 & (\textit{44\%}) & 35 & (\textit{42\%}) & 84\\
         10 heads & 39 & (\textit{44\%}) & 33 & (\textit{40\%}) & 83\\
         5 heads & 40 & (\textit{45\%}) & 36 & (\textit{42\%}) & 85\\
         3 heads & 52 & (\textit{58\%}) & 45 & (\textit{56\%}) & 81\\
         2 heads & 45 & (\textit{51\%}) & 36 & (\textit{46\%}) & 79\\
         1 heads & 45 & (\textit{51\%}) & 36 & (\textit{45\%}) & 80\\
         \hline
         \hline
        \end{tabularx}
    \caption{Stability of the topological classifier under attacks generated for the SST2 dataset. 
        \textbf{Avoided attacks} = number of attacks for which the result of the topological classifier does not change (89 attacks overall). 
        \textbf{Avoided Common Attacks} = consider only the attacks for which the topological classifier was initially correct.}
    \label{Tab:adversarial_attacks}
\end{table}

\cref{Tab:adversarial_attacks} shows that the topological model is much more stable than the BERT model. 
Only half of the attacks that succeed on BERT also succeed in fooling the topological classifier, which is surprising since the attention graphs are coming from the fooled BERT model.  
Even more, the stability of our model does not decrease with the number of considered heads. 
Even with persistence images coming from less than 5 heads, the adversarial attack efficiency exceeds slightly 50\%. 
This suggest that the robustness of the classifier based on persistence homology is not due to the large amount of input images.

\begin{figure}
    \centering
    \includegraphics[scale=0.25]{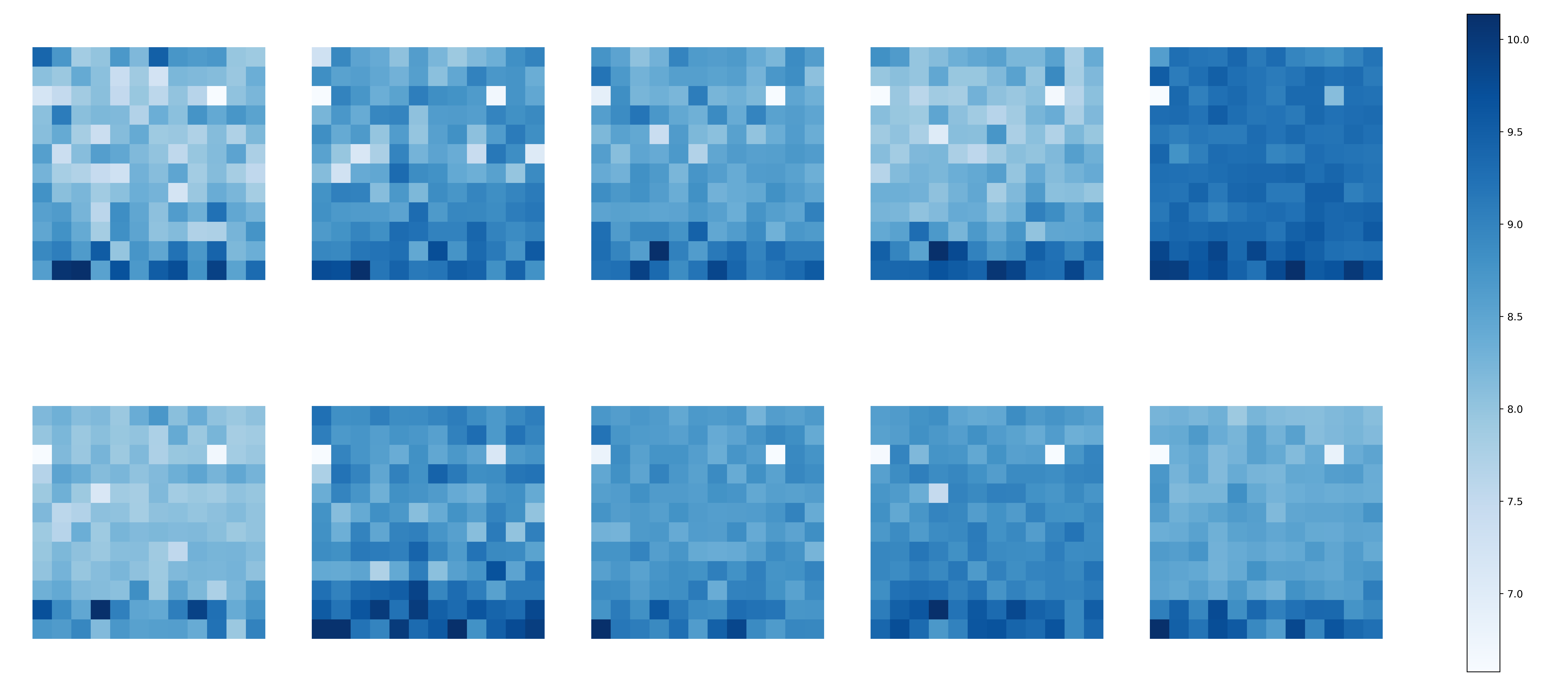}
    \caption{Perturbations of the persistence images for the first 10 attacks. 
    A pixel represents the head in the corresponding layer and column. 
    The darker the pixel, the larger the Euclidean distance between persistence images. The scale is logarithmic.}
    \label{fig:image_perturbation}
\end{figure}

Furthermore, the robustness is not due to the stability against adversarial attacks of the persistence images. 
\cref{fig:image_perturbation} displays the perturbation between the persistence images before and after the 10 first attacks. 
The squares represent the attention heads sorted by layers, with one pixel per head. The darker the pixel, the larger the Euclidean distance between the images generated from the head. 
We consider both the persistence images of dimension 0 and 1 by summing up their differences. 

Looking at the perturbation value of one head across different attacks, we notice that the perturbation value highly depends on the attack. 
For example, the perturbation value of the first head in the first layer (upper left corner) varies up to a factor of 100 between different attacks. 
In general, the images undergo modification before and after the attack.
But even if the images change, the model's output remains constant.

There are no canonical ways to analyze the perturbations in the attention maps. 
The sentence length can vary before and after the attack; therefore, the dimension of the attention maps can also vary. 
Hence, to measure the Euclidean distance, for example, one must first resize one of the attention maps to match the dimension of the other. 
Shrinking the largest is unsuitable as it removes the [SEP] column and will drastically change the attention graph's structure.
The same goes for the solution of padding the smaller attention map. 
This illustrates one advantage that persistence images adduce to interpretability methods: they allow to compare the model's behavior when it faces two sentences of different lengths.

It is worth noticing that the attacks do not produce correct or understandable movie reviews for SST2 (see \cref{tab:adversarial_attacks_example}) in general. 
The topological classifier seems to worry less about the overall meaning of the sentence. 
It might represent the input more abstractly and hence procure a stability towards meaning-switching words that is more appreciated for classification.

\begin{table}
    \centering
    \begin{tabularx}{1.0\textwidth}{>{\centering\arraybackslash}X |
            >{\centering\arraybackslash}X}
         \textbf{Before Attack} & \textbf{After Attack} \\
         \hline
         It's a \textcolor{ForestGreen}{charming} and often \textcolor{ForestGreen}{affecting} journey. $(1)$ & It 's a \textcolor{red}{cutie} and often \textcolor{red}{afflicts} journey. $(0)$ \\
         \hdashline
         Unflinchingly \textcolor{ForestGreen}{bleak} and desperate $(0)$ & Unflinchingly \textcolor{red}{eerie} and desperate $(1)$\\
         \hdashline
         Allows \textcolor{ForestGreen}{us} to hope that Nolan is poised to \textcolor{ForestGreen}{embark} a \textcolor{ForestGreen}{major} career as a commercial yet \textcolor{ForestGreen}{inventive filmmaker}. $(1)$ & 
         Allows \textcolor{red}{ourselves} to hope that Nolan is poised to \textcolor{red}{embarked} a \textcolor{red}{severe} career as a commercial yet \textcolor{red}{novelty superintendent}. $(0)$\\
         \hdashline
         The acting, costumes, music, cinematography and sound are all \textcolor{ForestGreen}{astounding} given the production's austere locales. $(1)$ & 
         The acting, costumes, music, cinematography and sound are all \textcolor{red}{breathless} given the production's austere locales. $(0)$ \\
         \hdashline
         It's slow -- \textcolor{ForestGreen}{very, very slow}. $(0)$ &
         It's slow -- \textcolor{red}{pretty, perfectly lent}. $(1)$\\
         \hline
    \end{tabularx}
    \caption{Five sample attacks generated by TextAttack for the SST2 dataset. 
    BERT classifies the reviews as positive $(1)$ or negative $(0)$.}
    \label{tab:adversarial_attacks_example}
\end{table}

The topological model presents greater stability than BERT. 
Furthermore, the stability is maintained regardless of the number of heads considered for the images and the impact of the attacks on the images. 
The stability must therefore be intrinsic to the classification itself. 

This robustness and high classification performance make our topological model more suitable than BERT when consistency and stability are needed.

\section{Conclusion}
In this work, we proposed numerous experiments on persistent homology applied for text classification. 
The model we present outperforms the baselines from BERT by $2\%$ and has higher robustness against adversarial attacks. 
We presented a new perspective on the specialization of BERT's attention heads using persistence diagrams, and also developed a new BERT attention head scoring technique.

Our most surprising finding is the efficiency of our proposed ratings, allowing us to consider only ten attention heads out of 144 with no reduction in accuracy on the test dataset or stability. 
Although the attention to the [SEP] token was assumed to have a \textit{no-op} behavior (\cite{what_does_bert_look_at}), a majority of the best scoring heads showcase this pattern, suggesting that through the lens of TDA, the attention to [SEP] displays valuable information for the classification task.

One possible direction for future research is to extend the tools from TDA to other types of NLP tasks. 
We recommend using ordinary persistence homology up to the first dimension to avoid computational complexity and to use more powerful vector representations than persistence images like the ones computed by the Persformer (\cite{persformer}). 
We also propose applying our rating approach to identify the most relevant heads and prune the others, which could increase performance. 
Lastly, we suggest training a specific classifier to detect adversarial textual attacks from the topology of the attention graphs.

\section{Acknowledgements}
This work was supported by the Swiss Innovation Agency (Innosuisse project 41665.1 IP-ICT).
We thank Matthias Kemper for helpful discussions and constructive comments on the paper.
\newline

\ifthenelse{\boolean{is_thesis}}{
For his patience every time my code crashed with no legitimate reason, for his meticulous help when proofreading my french-style English, and for his endless enthusiasm about the project, my thanks goes to Raphael Reinauer. I would also like to thank Milica Gasic who accepted to be the external expert for this thesis. I hope that my neophyte discovery will make you appreciate even more your domain of expertise. A special thanks goes to Virgile Constantin who was my precious colleague for this last semester in Master, always in the same working space and always cheerful. Finally my thanks goes to Kathryn Hess Bellwald whose answer everytime Raphael said \enquote{I will ask Kathryn} was a \enquote{yes}. 
}{}

\printbibliography
\newpage
\section*{Appendix}
\appendix

\addcontentsline{toc}{section}{Appendix}
\renewcommand{\thesubsection}{\Alph{subsection}}
\subsection{MultiDim UMAPs}\label{appendix_multi_dim_umap}
For this filtration, there is a clear cluster in the UMAPs of the 1-dim diagrams observed across the three sentences. 

\begin{figure}[H]
\begin{subfigure}{.33\textwidth}
  \centering
  \includegraphics[width=.95\linewidth]{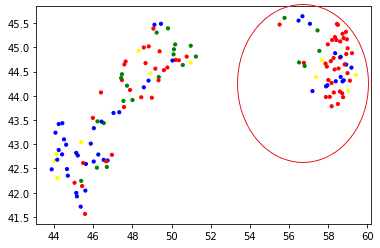}
  \caption{Sentence 1}
  \label{fig:sent_1_cola_twodim_1}
\end{subfigure}%
\begin{subfigure}{.33\textwidth}
  \centering
  \includegraphics[width=.95\linewidth]{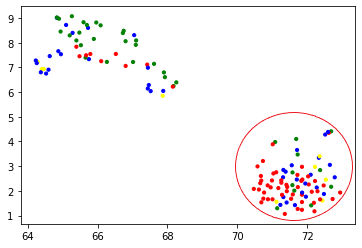}
  \caption{Sentence 2}
  \label{fig:sent_2_cola_twodim_1}
\end{subfigure}
\begin{subfigure}{.33\textwidth}
  \centering
  \includegraphics[width=.95\linewidth]{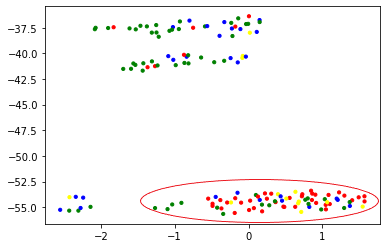}
  \caption{Sentence 3}
  \label{fig:sent_3_cola_twodim_1}
\end{subfigure}
\caption{UMAPs for the 1-features of MultiDim filtration. Clusters are circled in red. They contain 49, 86, and 72 elements for sentences 1, 2 and 3 respectively. Cluster 2 contains all elements of cluster 1. Cluster 3 shares 45 elements with it, and 58 with cluster 2.}
\label{fig:umap_cola_twodim_1}
\end{figure}

This time, they represent diagrams that contain almost only points on the diagonal $y=x$. 
Those points represent 1-persistence features that vanish at the time they are born. 
In other terms, these points represent a \enquote{triangular} cycle, a cycle formed by three edges only. 
When the third edge is added, a 2-cell is also added and fills the inside of the triangle, making the hole of the cycle disappear. 
As previously observed, the clusters across the sentences share globally the same heads. 

For the second homological dimension diagrams, a similar pattern is observed. 

\begin{figure}[H]
\begin{subfigure}{.33\textwidth}
  \centering
  \includegraphics[width=.95\linewidth]{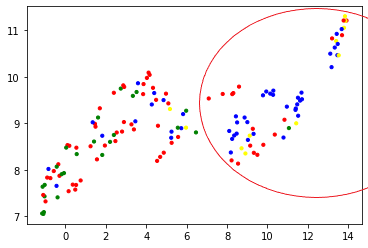}
  \caption{Sentence 1}
  \label{fig:sent_1_cola_twodim_2}
\end{subfigure}%
\begin{subfigure}{.33\textwidth}
  \centering
  \includegraphics[width=.95\linewidth]{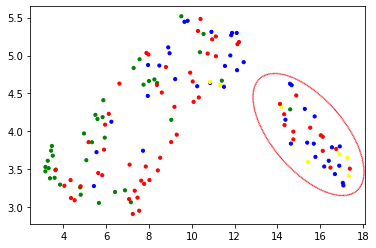}
  \caption{Sentence 2}
  \label{fig:sent_2_cola_twodim_2}
\end{subfigure}
\begin{subfigure}{.33\textwidth}
  \centering
  \includegraphics[width=.95\linewidth]{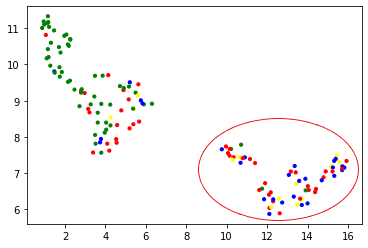}
  \caption{Sentence 3}
  \label{fig:sent_3_cola_twodim_2}
\end{subfigure}
\caption{UMAPs for the 2-features of the MultiDim filtration. Clusters are circled in red. They contain 61, 38, and 64 elements for sentences 1, 2 and 3 respectively. Cluster 2 contains 36 elements with cluster 1. Cluster 3 shares 46 elements with it, and 33 with cluster 2.}
\label{fig:umap_cola_twodim_2}
\end{figure}

The difference between the diagrams that are inside the cluster from the ones that are outside is that the former have only 2-holes with a high birth-time and the latter have 2-holes with varying birth-times.
When transformed into persistence images, the diagrams outside the cluster will display a richer variety of patterns, compared to the diagrams inside the cluster whose images are similar: high value pixels on the top right corner, and small values everywhere else. 
Again, the clusters across sentences share a similar composition.

\subsection{Directed UMAPs}\label{appendix_directed_umap}
We observe similar clusters as in the MultiDim filtration case. 

\begin{figure}[H]
\begin{subfigure}{.33\textwidth}
  \centering
  \includegraphics[width=.95\linewidth]{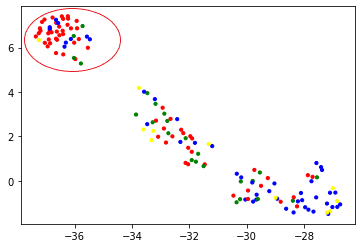}
  \caption{Sentence 1}
  \label{fig:sent_1_cola_directed_1}
\end{subfigure}%
\begin{subfigure}{.33\textwidth}
  \centering
  \includegraphics[width=.95\linewidth]{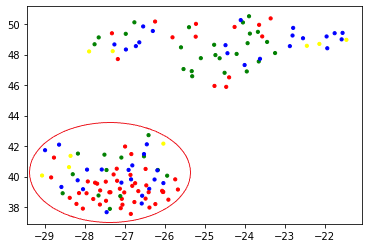}
  \caption{Sentence 2}
  \label{fig:sent_2_cola_directed_1}
\end{subfigure}
\begin{subfigure}{.33\textwidth}
  \centering
  \includegraphics[width=.95\linewidth]{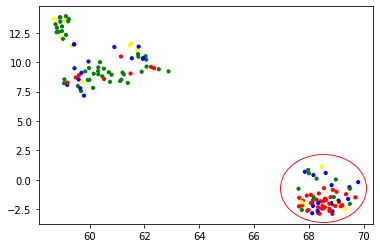}
  \caption{Sentence 3}
  \label{fig:sent_3_cola_directed_1}
\end{subfigure}
\caption{UMAPs for the 1-features of the Directed filtration. Clusters are circled in red. They contain 51, 82, and 68 elements for sentences 1, 2 and 3 respectively. Cluster 2 shares 41 elements with cluster 1. Cluster 3 shares 35 elements with it, and 50 with cluster 2.}
\label{fig:umap_cola_directed_1}
\end{figure}

\begin{figure}[H]
\begin{subfigure}{.33\textwidth}
  \centering
  \includegraphics[width=.95\linewidth]{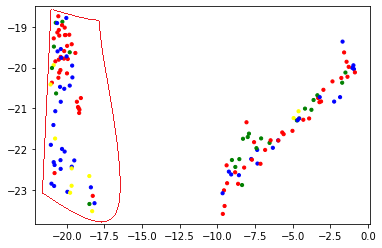}
  \caption{Sentence 1}
  \label{fig:sent_1_cola_directed_2}
\end{subfigure}%
\begin{subfigure}{.33\textwidth}
  \centering
  \includegraphics[width=.95\linewidth]{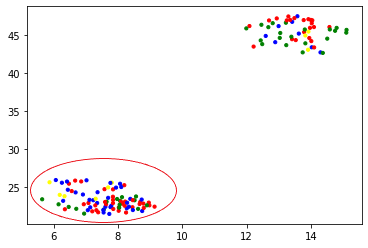}
  \caption{Sentence 2}
  \label{fig:sent_2_cola_directed_2}
\end{subfigure}
\begin{subfigure}{.33\textwidth}
  \centering
  \includegraphics[width=.95\linewidth]{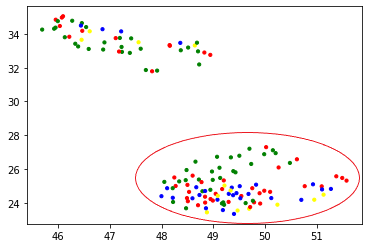}
  \caption{Sentence 3}
  \label{fig:sent_3_cola_directed_2}
\end{subfigure}
\caption{UMAPs for the 2-features of the Directed filtration. Clusters are circled in red. They contain 63, 91, and 96 elements for sentences 1, 2 and 3 respectively. Cluster 2 shares 58 elements with cluster 1. Cluster 3 shares 60 elements with it, and 79 with cluster 2}
\label{fig:umap_cola_directed_2}
\end{figure}

Their meaning is identical: diagrams with almost only diagonal points for the first-dimensional features and top right points for the second-dimensional features.
Again the composition of the clusters is similar across sentences. 

\subsection{Pruning heads across models and datasets}\label{appendix_pruning}
We further investigate how the our pruning procedure behaves across different dataset and across different fine-tuned BERT models.
\begin{table}[H]
    \centering
    \begin{tabularx}{1.0\textwidth} { |
            >{\raggedright\arraybackslash}X :
            >{\raggedleft\arraybackslash}X :
            >{\raggedleft\arraybackslash}X :
            >{\raggedleft\arraybackslash}X :
            >{\raggedleft\arraybackslash}X |}
         \hline
         \hline
          & \textbf{\thead{CoLA \\ 4 Epochs \\ Ordinary \\ Max}} & \textbf{\thead{IMDB \\ 4 Epochs \\ Ordinary \\ Max}} & \textbf{\thead{SPAM \\ 4 Epochs \\ Ordinary \\ Max}} & \textbf{\thead{SST2 \\ 4 Epochs \\ Ordinary \\ Max}} \\
          \hline
          \textbf{BERT} & $0.518~/~80.6$ & $\mathbf{0.843~/~92.2}$ & $\underline{0.993~/~99.8}$ & $\mathbf{0.853~/~92.7}$ \\
          \hline
          144 Heads & $0.539~/~81.2$ & $\underline{0.776~/~88.8}$ & $\mathbf{0.999~/~99.9}$ & $\underline{0.799~/~89.8}$ \\
          70 Heads & $0.557~/~81.3$ & $0.664~/~83.0$ & $\underline{0.993~/~99.8}$ & $0.784~/~89.2$ \\   
          50 Heads & $0.557~/~81.2$ & $0.611~/~79.1$ & $0.957~/~98.9$ & $0.782~/~89.1$ \\   
          30 Heads & $\mathbf{0.582~/~82.4}$ & $0.596~/~79.1$ & $\underline{0.993~/~99.8}$ & $0.771~/~88.5$ \\
          10 Heads & $0.545~/~81.0$ & $0.413~/~70.7$ & $0.993~/~99.5$ & $0.766~/~88.3$ \\
          5 Heads & $0.553~/~81.7$ & $0.412~/~70.1$ & $0.978~/~99.5$ & $0.459~/~72.4$ \\
          3 Heads & $0.554~/~81.9$ & $0.395~/~65.4$ & $0.948~/~98.6$ & $0.460~/~72.7$ \\
          2 Heads & $\underline{0.560~/~82.1}$ & $0.231~/~60.6$ & $0.710~/~93.6$ & $0.307~/~63.5$ \\
          1 Head & $0.492~/~79.4$ & $0.184~/~57.4$ & $0.396~/~88.0$ & $-0.026~/~48.6$ \\
          \hline
          \hline
    \end{tabularx}
    \caption{Performances of the topological models from the different tasks while considering different amounts of high scoring heads. These are determined from the base PI-dataset, and \textbf{the evaluated model is specific to CoLA.}}
    \label{Tab:table_best_heads_results_tasks}
\end{table}

\begin{table}[H]
    \centering
    \begin{tabularx}{1.0\textwidth} { |
            >{\raggedright\arraybackslash}X :
            >{\raggedleft\arraybackslash}X :
            >{\raggedleft\arraybackslash}X :
            >{\raggedleft\arraybackslash}X :
            >{\raggedleft\arraybackslash}X |}
         \hline
         \hline
          & \textbf{\thead{CoLA \\ 4 Epochs \\ Ordinary \\ Max}} & \textbf{\thead{IMDB \\ 4 Epochs \\ Ordinary \\ Max}} & \textbf{\thead{SPAM \\ 4 Epochs \\ Ordinary \\ Max}} & \textbf{\thead{SST2 \\ 4 Epochs \\ Ordinary \\ Max}} \\
          \hline
          \textbf{BERT} & $0.518~/~80.6$ & $\mathbf{0.843~/~92.2}$ & $0.993~/~99.8$ & $\mathbf{0.853~/~92.7}$ \\
          \hline
          144 Heads & $0.539~/~81.2$ & $\underline{0.787~/~89.4}$ & $\underline{0.999~/~99.9}$ & $0.799~/~89.8$ \\
          70 Heads & $0.557~/~81.3$ & $0.777~/~89.0$ & $0.993~/~99.8$ & $0.780~/~89.0$ \\   
          50 Heads & $0.557~/~81.2$ & $0.762~/~88.2$ & $0.993~/~99.8$ & $0.767~/~88.3$ \\   
          30 Heads & $\mathbf{0.582~/~82.4}$ & $0.783~/~89.3$ & $\mathbf{1.0~/~100}$ & $\underline{0.798~/~89.9}$ \\
          10 Heads & $0.545~/~81.0$ & $0.779~/~89.0$ & $0.993~/~99.8$ & $0.778~/~88.9$ \\
          5 Heads & $0.553~/~81.7$ & $0.770~/~88.6$ & $0.836~/~95.5$ & $0.789~/~89.5$ \\
          3 Heads & $0.554~/~81.9$ & $0.767~/~88.5$ & $0.831~/~95.5$ & $0.745~/~87.3$ \\
          2 Heads & $\underline{0.560~/~82.1}$ & $0.695~/~84.8$ & $0.786~/~94.4$ & $0.717~/~85.8$ \\
          1 Head & $0.492~/~79.4$ & $0.560~/~77.2$ & $0.772~/~94.3$ & $0.721~/~85.9$ \\
          \hline
          \hline
    \end{tabularx}
    \caption{Performances of the topological models from the different tasks while considering different amounts of high scoring heads. These are determined for each PI-datasets, and \textbf{the evaluated model is specific to the current task.}}
    \label{Tab:table_best_heads_results_tasks_specific}
\end{table}

For the head scores from CoLA, the performance of the model on the other tasks decreases with decreasing number of considered images. 
But when the head scores are determined for each PI-dataset, the performance remains constant when at least 10 heads are considered, and decreases slowly when less than 10 heads are considered. 
For the SPAM dataset, we even observe a perfect score of 100\% accuracy obtained while considering 30 heads.
The boosting effect of pruning images are the most significant for the CoLA dataset.

\begin{table}[H]
    \centering
    \begin{tabularx}{1.0\textwidth} { |
            >{\raggedright\arraybackslash}X :
            >{\raggedleft\arraybackslash}X :
            >{\raggedleft\arraybackslash}X :
            >{\raggedleft\arraybackslash}X |}
         \hline
         \hline
          & \textbf{\thead{CoLA \\ 4 Epochs \\ Ordinary \\ Max}} & \textbf{\thead{CoLA \\ 4 Epochs \\ MultiDim \\ Max}} & \textbf{\thead{CoLA \\ 4 Epochs \\ Directed \\ Max}} \\
          \hline
          \textbf{BERT} & $0.518~/~80.6$ & $0.518~/~80.6$ & $0.518~/~80.6$ \\
          \hline
          144 Heads & $0.539~/~81.2$ & $0.552~/~81.5$ & $0.532~/~80.7$ \\
          70 Heads & $0.557~/~81.3$ & $0.544~/~81.2$ & $\mathbf{0.551~/~81.8}$ \\   
          50 Heads & $0.557~/~81.2$ & $0.546~/~81.1$ & $0.547~/~81.6$ \\   
          30 Heads & $\mathbf{0.582~/~82.4}$ & $\mathbf{0.561~/~82.0}$ & $\underline{0.548~/~81.6}$ \\
          10 Heads & $0.545~/~81.0$ & $0.555~/~81.5$ & $0.538~/~81.2$ \\
          5 Heads & $0.553~/~81.7$ & $0.557~/~81.6$ & $0.528~/~80.9$ \\
          3 Heads & $0.554~/~81.9$ & $0.554~/~81.6$ & $0.538~/~81.0$ \\
          2 Heads & $\underline{0.560~/~82.1}$ & $\underline{0.556~/~81.7}$ & $0.503~/~79.7$ \\
          1 Head & $0.492~/~79.4$ & $0.485~/~79.2$ & $0.455~/~78.1$ \\
          \hline
          \hline
    \end{tabularx}
    \caption{Performances of the topological models for different types of persistent homology for the CoLA dataset while considering different amounts of high scoring heads. These are determined specifically for each PI-dataset.}
    \label{tab:table_best_heads_results_filtrations}
\end{table}

Pruning the images is beneficial for performance, with a more significant effect for the Ordinary and Directed filtrations. 
Without pruning, the MultiDim filtration outperforms the others, but the ordinary persistence combined with pruning reaches the same peak performance of 82\% in accuracy. 
Interestingly, to increase the performance of the topological classifier, one should prefer to remove some well-chosen input images, rather than considering more complex and computation-demanding filtrations.

\cref{Tab:table_best_heads_results_inverted} shows the result for different symmetry functions and a different number of fine-tuning epochs. 
Here, the line \textit{10 heads} corresponds to the performance obtained by the model while trained on images from 134 attention heads (we removed the images from the 10 best heads).

\begin{table}[H]
    \centering
    \begin{tabularx}{1.0\textwidth} { |
            >{\raggedright\arraybackslash}X :
            >{\raggedleft\arraybackslash}X :
            >{\raggedleft\arraybackslash}X :
            >{\raggedleft\arraybackslash}X :
            >{\raggedleft\arraybackslash}X :
            >{\raggedleft\arraybackslash}X |}
         \hline
         \hline
          & \textbf{\thead{CoLA \\ 4 Epochs \\ Ordinary \\ Max}} & \textbf{\thead{CoLA \\ 4 Epochs \\ Ordinary \\ Min}} & \textbf{\thead{CoLA \\ 4 Epochs \\ Ordinary \\ Mean}} & \textbf{\thead{CoLA \\ 10 Epochs \\ Ordinary \\ Max}} & \textbf{\thead{CoLA \\ 20 Epochs \\ Ordinary \\ Max}}\\
          \hline
          \textbf{BERT} & $0.518~/~80.6$ & $0.518~/~80.6$ & $0.518~/~80.6$ & $0.561~/~82.2$ & $0.591~/~83.3$\\
          \hline
          70 Heads & $0.0~/~30.9$ & $0.540~/~81.2$ & $0.0~/~30.9$ & $0.562~/~81.9$ & $0.0~/~30.9$\\   
          50 Heads & $0.486~/~79.4$ & $0.505~/~80.1$ & $0.492~/~15.9$ & $0.564~/~82.2$ & $\underline{0.593~/~83.4}$\\   
          30 Heads & $0.517~/~80.4$ & $\mathbf{0.559~/~82.0}$ & $0.499~/~79.6$ & $0.565~/~82.4$ & $0.573~/~82.7$\\
          10 Heads & $0.533~/~80.9$ & $\underline{0.538~/~81.3}$ & $0.484~/~76.6$ & $0.574~/~82.7$ & $0.584~/~83.0$\\
          5 Heads & $0.514~/~80.4$ & $0.536~/~81.2$ & $\mathbf{0.545~/~81.6}$ & $0.561~/~82.0$ & $0.583~/~83.0$\\
          3 Heads & $\underline{0.537~/~80.9}$ & $0.531~/~81.1$ & $\underline{0.531~/~81.1}$ & $\underline{0.574~/~82.7}$ & $0.573~/~82.6$\\
          2 Heads & $0.540~/~80.7$ & $0.546~/~80.9$ & $0.526~/~80.9$ & $0.585~/~82.5$ & $\mathbf{0.593~/~83.4}$\\
          1 Head & $\mathbf{0.537~/~81.0}$ & $0.525~/~80.8$ & $0.538~/~80.3$ & $\mathbf{0.584~/~82.7}$ & $0.566~/~82.4$\\
          \hline
          \hline
    \end{tabularx}
    \caption{Performances of the topological models while removing different numbers of high scoring heads. For examples, \textit{5 heads} means that we removed the images generated by the 5 best heads, always determined from the base model.}
    \label{Tab:table_best_heads_results_inverted}
\end{table}

The general tendency is a constant performance while the number of considered images increases. 
The exceptions are for the \textit{mean} symmetry function where a neat increase in accuracy occurs, and the low accuracy when only half of the worst performing heads are considered. 

\newpage
\subsection{Data and model specifications}

\vspace{5cm}
\begin{table}[h]
    \centering
    \ra{1.5}
    \begin{tabular}{c|c|c|c|c|c}
    \hline
    \hline
    \thead{Type of persistence \\ homology} & \thead{Feature \\ dimension} & Picture frame & Resolution & Library & Standardize\\
     \hline
     Ordinary & 0 & $[0, 0.01] \times [0, 1]$ & $5 \times 50$ & Gudhi & -\\
     Ordinary & 1 & $[0, 1] \times [0.99, 1]$ & $50 \times 5$ & Gudhi & rotation of 45°\\
     \hdashline
    MultiDim & 0 & $[0, 0.01] \times [0, 1]$ & $5 \times 50$ & Gudhi & \thead{padding \\with zeros to \\$50\times50$}\\
    MultiDim & 1 & $[0.5, 1] \times [0.5, 1]$ & $50 \times 50$ & Gudhi & -\\
    MultiDim & 2 & $[0.7, 1] \times [0.999, 1]$ & $50 \times 5$ & Gudhi & \thead{padding \\with zeros to \\$50\times50$}\\
    \hdashline
    Directed & 0,1,2 & $[0, 0.01] \times [0, 1]$ & $30 \times 30$ & Giotto-tda & -\\
    \end{tabular}
    \caption{Parameters for the persistence image computation per type of persistent homology considered. 
    Various picture frames and resolutions are considered to use to good advantage the particular structure of the persistence diagrams. 
    \textbf{Picture frame} = transformed part of the persistent diagram contained in $[0,1]\times [0,1]$. 
    \textbf{Standardize} = transformation to apply to the image to have identical image dimensions for each feature dimension.
    References for libraries: Gudhi (\cite{gudhi}), Giotto-tda (\cite{giotto-tda}).}
    \label{tab:parameters_persistence_images}
\end{table}
\vfil

\begin{table}[p]
    \centering
    \ra{2}
    \begin{tabular}{c|c|c}
    \hline
    \hline
    \thead{Type of persistence \\ homology} & Weight function & Motivation\\
     \hline
     Ordinary - 0 & $w\colon (x,y) \mapsto 1$ & -\\
     Ordinary - 1 & 
     $w\colon (x,y) \mapsto 
     \begin{cases}
        5\cdot (1-x), ~\text{if}~ x\geq 0.8\\
        1, ~\text{otherwise}.\\
    \end{cases}$ & \thead{decrease influence of\\ end-filtration features}\\
    \hdashline
    MultiDim - 0 & $w\colon (x,y) \mapsto 1$ & -\\
    MultiDim - 1 &
    $w\colon (x,y) \mapsto 
    \begin{cases}
        2 - \frac{x + y}{2} - e^{-10\cdot |x-y|}, ~\text{if}~ \frac{x+y}{2}\geq 0.9\\
        1.1 - e^{-10\cdot |x-y|}, ~\text{otherwise}.\\
    \end{cases}$ & \thead{decrease influence of\\ end-filtration and \\near diagonal features}\\
    MultiDim - 2 & 
    $w\colon (x,y) \mapsto
    \begin{cases}
        10\cdot (1-x), ~\text{if}~ x\geq 0.9\\
        1, ~\text{otherwise}.\\
    \end{cases}$ & \thead{decrease influence of\\ end-filtration features}\\
    \hdashline
    Directed - 0,1,2 & $w\colon (x,y) \mapsto 1$ & -\\
    \end{tabular}
    \caption{Weight functions we considered for the persistence image computation. 
    The core motivation is to decrease the importance of features appearing at the very end of the filtration, and of features really close to the diagonal $x=y$.}
    \label{tab:weight_function_persistence_images}
\end{table}

\begin{table}[p]
    \centering
    \begin{tabular}{>{\centering\arraybackslash} m{3cm} |
                    >{\centering\arraybackslash} m{12cm}}
        \hline
        \hline
        \thead{Type of persistence \\ homology} & Examples \\
        \hline
        Ordinary - 0 & \includegraphics[width=1.0\linewidth]{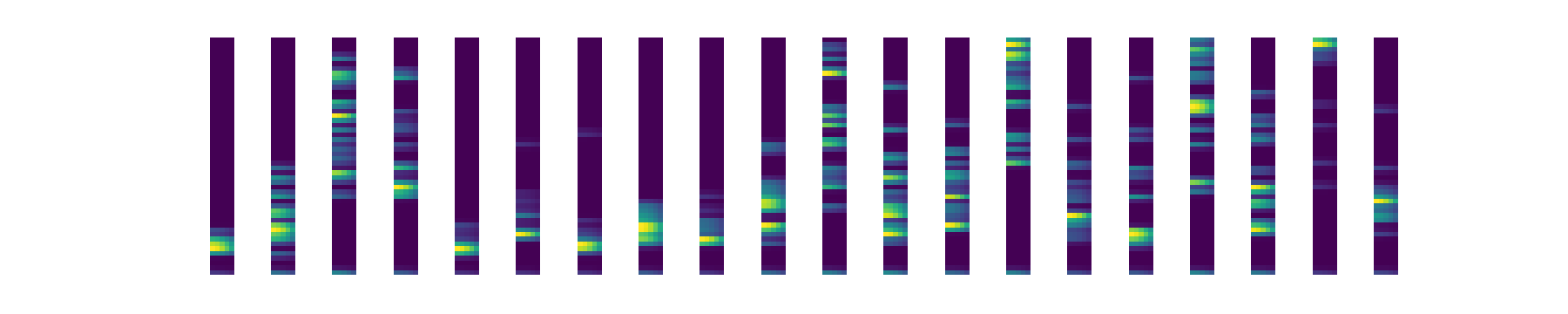}\\
        Ordinary - 1 & \includegraphics[width=1.0\linewidth]{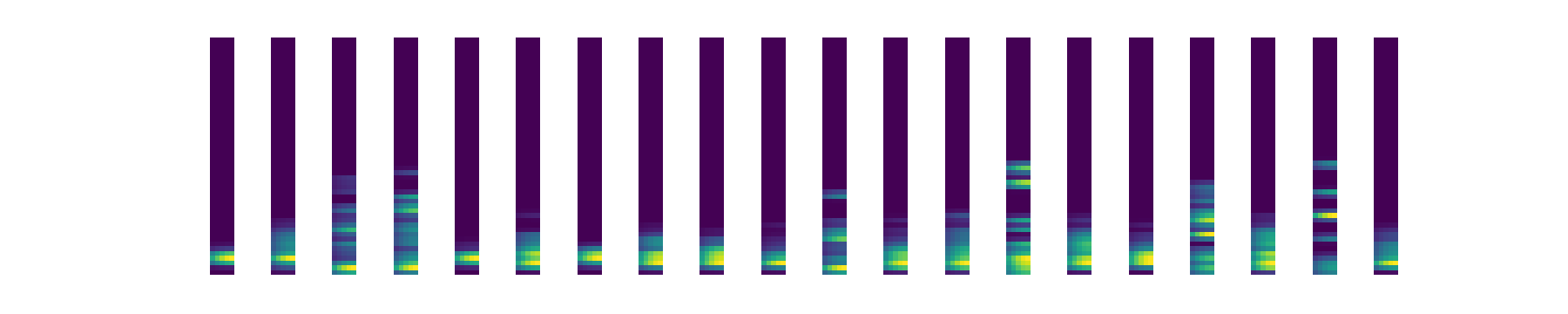}\\
        \hdashline
        MultiDim - 0 & \includegraphics[width=1.0\linewidth]{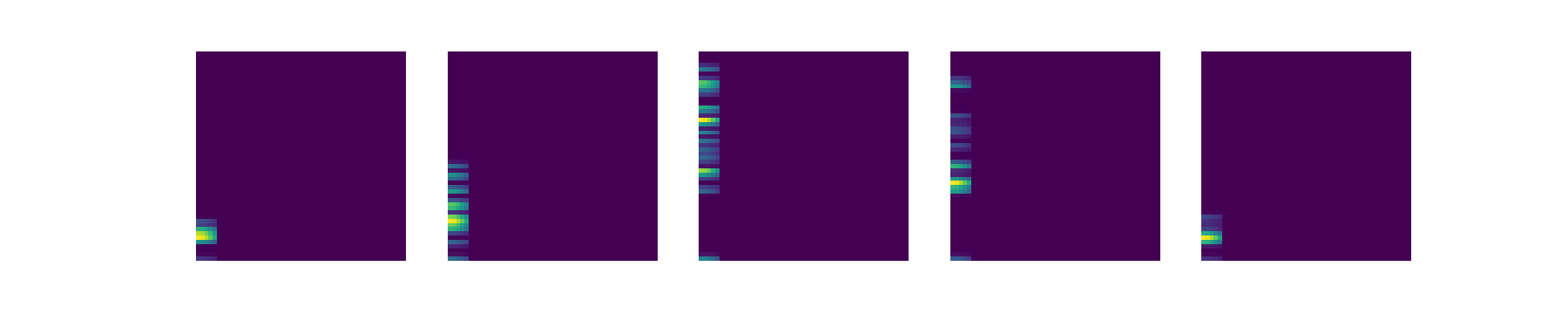}\\
        MultiDim - 1 & \includegraphics[width=1.0\linewidth]{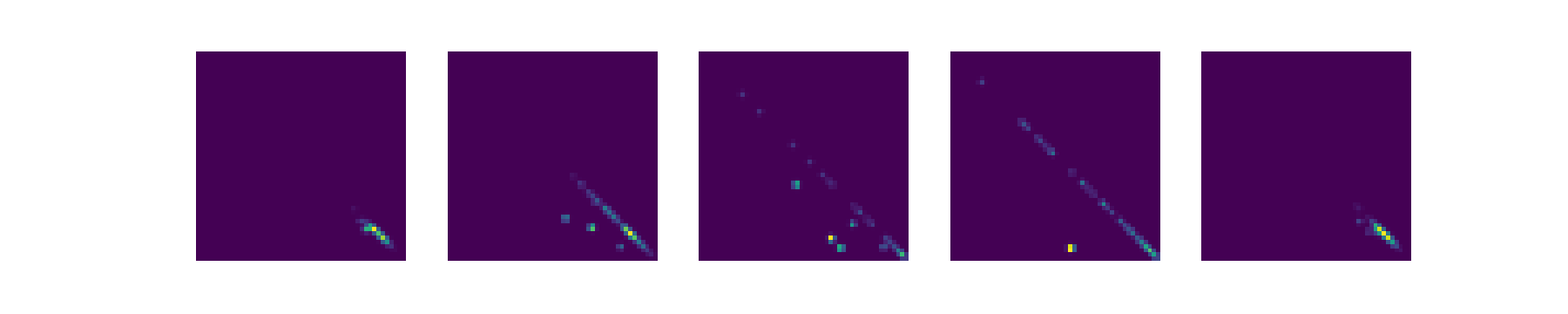}\\
        MultiDim - 2 & \includegraphics[width=1.0\linewidth]{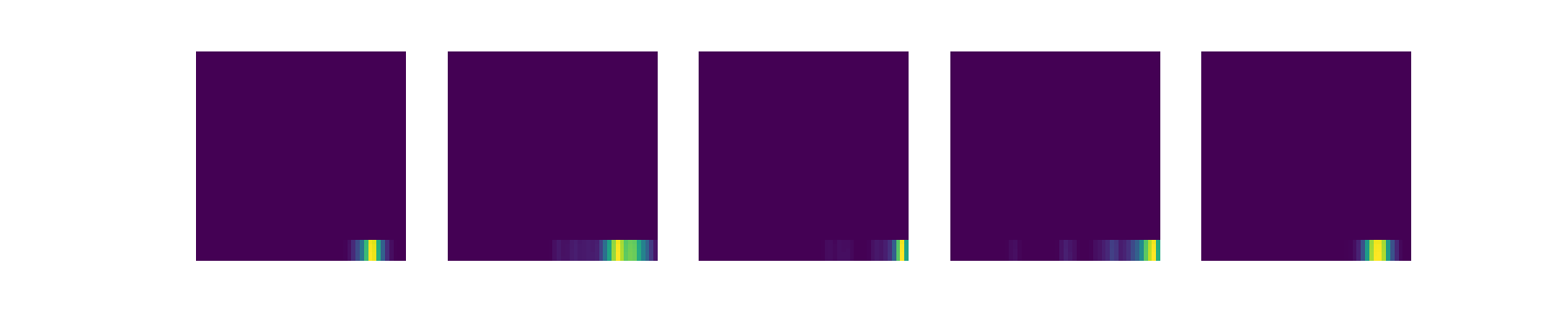}\\
        \hdashline
        Directed - 0 & \includegraphics[width=1.0\linewidth]{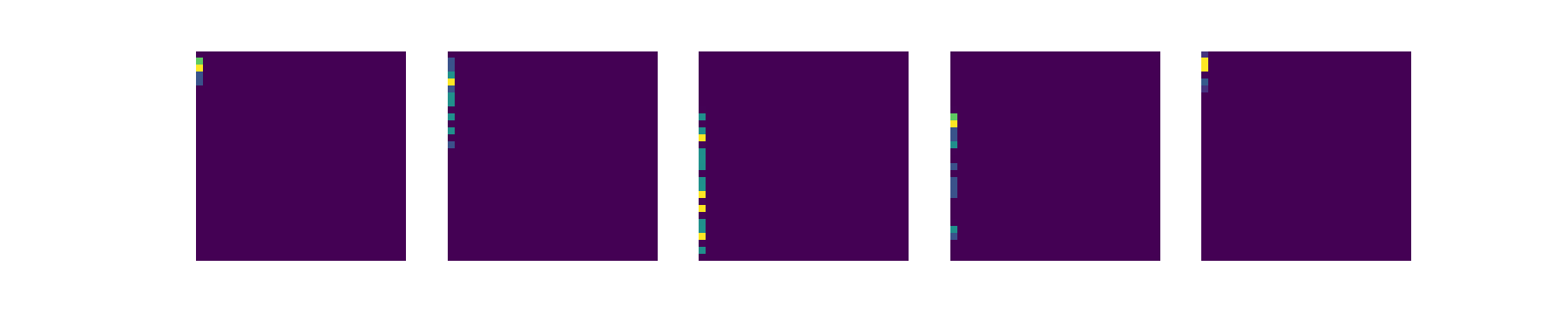}\\
        Directed - 1 & \includegraphics[width=1.0\linewidth]{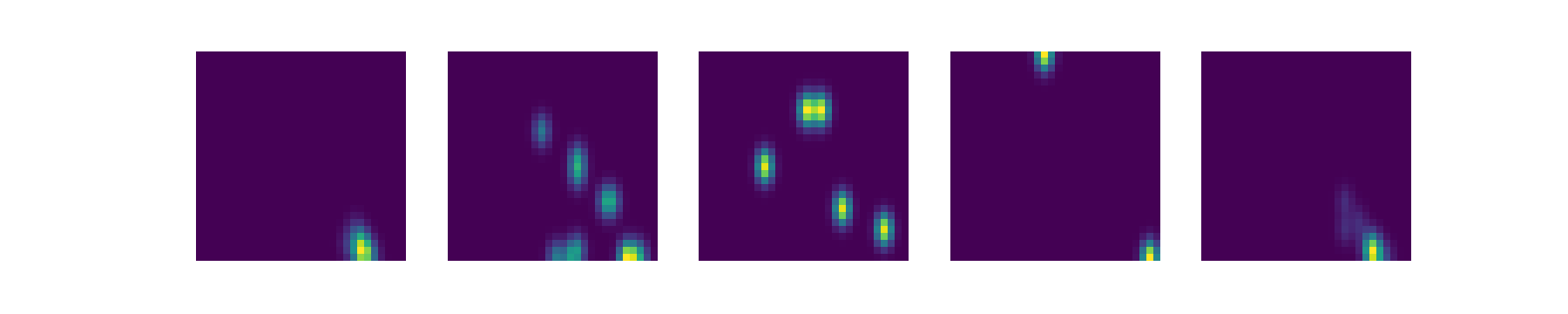}\\
        Directed - 2 & \includegraphics[width=1.0\linewidth]{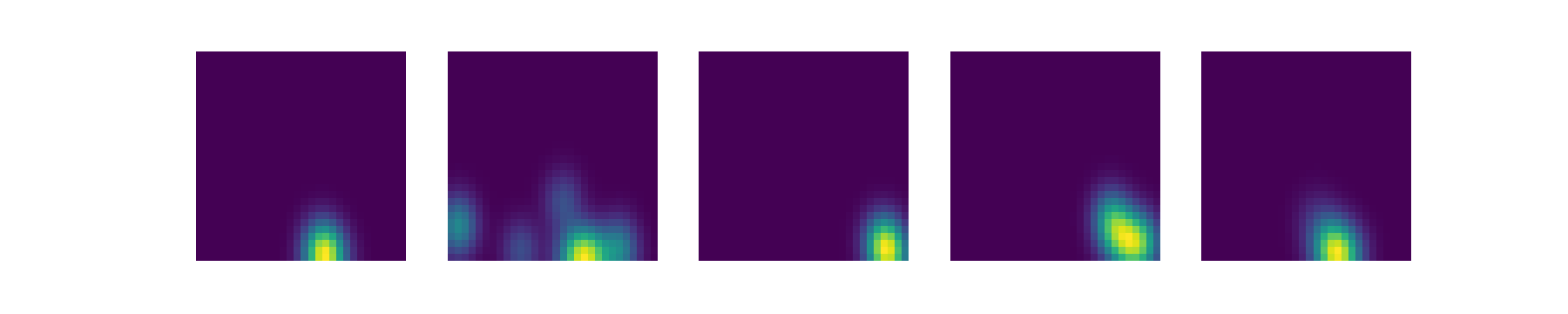}\\
    \end{tabular}
    \caption{Examples of persistence images for each type of persistent homology and each features dimension. 
    Notice that the persistence images are the horizontal reflection of the persistence diagrams.}
    \label{tab:examples_persistence_images_for_all_type}
\end{table}

\begin{landscape}

\begin{table}[p]
    \centering
    \ra{2}
    \small
    \begin{tabular}{c:c|c:c|c:c:c:c:c|c:c|c}
        \hline
        \hline
        Dataset & \thead{Type of persistence \\ homology} & Optimizer & lr & \# conv. & filters & dim. ker. & stride & pad & \# liner & dim. layer & dropout rate\\
        \hline
        CoLA & Ordinary & Adam & 6.7e-4 & 3 & $[15;25;45]$ & None & None & None & 1 & $[150]$ & 0.25\\
        CoLA & MultiDim & RMSprop & 7.5e-5 & 2 & $[15;25]$ & $[3;2]$ & $[3;2]$ & $[1;1]$ & 2 & $[500;120]$ & 0.25\\
        CoLA & Directed & RMSprop & 9.2e-4 & 3 & $[4;30;20]$ & $[2;1;2]$ & $[1;1;2]$ & $[1;0;1]$ & 3 & $[700;350;750]$ & 0.15\\
        \hdashline
        IMDB & Ordinary & Adam & 8.4e-5 & 2 & $[35;30]$ & None & None & None & 1 & $[180]$ & 0.25\\
        IMDB & MultiDim & Adam & 1.2e-3 & 3 & $[17;17;17]$ & $[2;3;2]$ & $[2;1;2]$ & $[1;0;0]$ & 3 & $[700;660;800]$ & 0.2\\
        \hdashline
        SPAM & Ordinary & Adam & 2.2e-4 & 2 & $[15;25]$ & None & None & None & 1 & $[700]$ & 0.2\\
        SPAM & MultiDim & Adam & 7e-4 & 3 & $[33;5;32]$ & $[1;2;1]$ & $[1;2;1]$ & $[0;1;0]$ & 2 & $[480;220]$ & 0.3\\
        \hdashline
        SST2 & Ordinary & Adam & 5e-4 & 1 & $[35]$ & None & None & None & 2 & $[190;940]$ & 0.25\\
        SST2 & MultiDim & Adam & 2.4e-5 & 2 & $[20;20]$ & $[2;1]$ & $[2;1]$ & $[1;0]$ & 2 & $[650;680]$ & 0.25\\
    \end{tabular}
    \caption{Architectures of all the topological classifiers we design, based on 
    running a hyperparameter search using the Optuna library (\cite{optuna}) for 500 trials. 
    \textbf{Dataset}= dataset on which the hyperparameter search was run. \textbf{Type of persistent homology} = type of the persistence images as input.
    The input is a tensor of shape $[288,50,5]$ for Ordinary, of shape $[432,50,50]$ for MultiDim, and of shape $[432,30,30]$ for Directed.
    \textbf{dim. ker / stride / pad} = dimension of the max pooling layers, None if no pooling.
    The dropout layer is right after the ReLU activation.}
    \label{tab:architecture_of_topological_models}
\end{table}

\end{landscape}

\end{document}